\documentclass[lettersize,journal]{IEEEtran}
    \usepackage{amsmath,amsfonts,amssymb}
    \usepackage{algorithmic}
    \usepackage{algorithm}
    \usepackage{array}
    \usepackage{xpatch}
    \usepackage{multirow}
    \usepackage{booktabs}
    \usepackage[caption=false,font=normalsize,labelfont=sf,textfont=sf]{subfig}
    \usepackage{textcomp}
    \usepackage{stfloats}
    \usepackage{url}
    \usepackage{verbatim}
    \usepackage{graphicx}
    \usepackage{cite}
    \usepackage{balance}
    \usepackage{bbm}
    \usepackage{color,xcolor}
    \usepackage{ulem}
    \usepackage{tikz}

    \renewcommand{\emph}{\textit}
    
    \def\mode{submission} % To disabthe colours/marks, modify the \mode to "submission" instead of "editing".
    \def\editing{editing}

\def\S{\mathcal{S}}
\def\F{\mathcal{F}}

\def\oneif{\mathbbm{1}}
\def\vec{\boldsymbol{v}}
\def\vrow{v^{\text{row}}}
\def\vcol{v^{\text{col}}}
\def\f{\boldsymbol{\phi}}
\ifx\mode\editing

    \newcommand{\revised}[1]{\textcolor{blue}{#1}}

  \newenvironment{summar}{\color{pink!75!blue}}{\color{black}}
\newenvironment{needreview}{\color{red}}{\color{black}}
\newenvironment{revisedparas}{\color{blue}}{\color{black}}
\newenvironment{revisingparas}{\color{orange}}{\color{black}}

\else

    \newcommand{\revised}[1]{#1}

\newenvironment{revisedparas}{}{}

\fi

\def\E{\mathbb{E}}

\makeatletter
\ExplSyntaxOn
\cs_new:Npn \bibColoredItems #1#2
  {
    \clist_map_inline:nn {#2} { \cs_new:cpn {bib@colored@##1} {#1} } 
  }
\ExplSyntaxOff

\newcommand\bib@setcolor[1]{%
  \ifcsname bib@colored@#1\endcsname
    \expandafter\color\expandafter{\csname bib@colored@#1\endcsname}
  \else
    \normalcolor
  \fi
}

\xpatchcmd\@bibitem
  {\item}
  {\bib@setcolor{#1}\item}
  {}{\PatchFailed}

\xpatchcmd\@lbibitem
  {\item}
  {\bib@setcolor{#2}\item}
  {}{\PatchFailed}
\makeatother

\bibColoredItems{blue}{}
\begin{document}
    
    \title{Measuring Diversity of Game Scenarios}%: A Survey}
    \author{\IEEEauthorblockN{Yuchen Li\IEEEauthorrefmark{1}\IEEEauthorrefmark{2},~\IEEEmembership{Student Member,~IEEE},
    Ziqi Wang\IEEEauthorrefmark{1},~\IEEEmembership{Student Member,~IEEE}, Qingquan Zhang\IEEEauthorrefmark{1},~\IEEEmembership{Member,~IEEE}, Bo Yuan\IEEEauthorrefmark{1},~\IEEEmembership{Member,~IEEE}
    and
    Jialin Liu\IEEEauthorrefmark{1}\IEEEauthorrefmark{3},~\IEEEmembership{Senior Member,~IEEE}}
    \thanks{\IEEEauthorrefmark{1}Guangdong Provincial Key Laboratory of Brain-inspired Intelligent Computation, Department of Computer Science and Engineering, Southern University of Science and Technology, Shenzhen, China.}
    \thanks{\IEEEauthorrefmark{2}Tandon School of Engineering, New York University, USA.}
    \thanks{\IEEEauthorrefmark{3}School of Data Science, Lingnan University, Hong Kong SAR, China.}
    \thanks{Y. Li and Z. Wang contributed equally to this work.}
    }

    \def\removed{
   \author{Yuchen Li,~\IEEEmembership{Student Member,~IEEE}, Ziqi Wang,~\IEEEmembership{Student Member,~IEEE}, Qingquan Zhang,~\IEEEmembership{Member,~IEEE}, Bo Yuan,~\IEEEmembership{Member,~IEEE},
    Xin Wang, 
    Jialin Liu,~\IEEEmembership{Senior Member,~IEEE}
    % <-this % stops a space
    %\thanks{Y. Li, Z. Wang, Q. Zhang, B. Yuan and J. Liu are with the Research Institute of Trustworthy Autonomous System, Southern University of Science and Technology (SUSTech), Shenzhen, China.}
    \thanks{Y. Li, Z. Wang, Q. Zhang, B. Yuan and J. Liu are with the Department of Computer Science and Engineering of SUSTech.}
    \thanks{Y. Li is now with New York University, USA.}
    \thanks{Z. Wang is now with Hong Kong University of Science and Technology, Hong Kong SAR, China.}
    \thanks{J. Liu is now with Lingnan University, Hong Kong SAR, China.}
% Corresponding author: Jialin Liu (liujl@sustech.edu.cn).
    \thanks{Y. Li and Z. Wang contributed equally to this work.}
}}
    
    
    %\tableofcontents
    \maketitle
    
    \begin{abstract}
    \revised{This survey comprehensively reviews the metrics for measuring the diversity of game scenarios, spotlighting the innovative use of procedural content generation and other fields as cornerstones for enriching player experiences through diverse game scenarios.} By traversing a wide array of disciplines, from affective modeling and multi-agent systems to psychological studies, our research underscores the importance of diverse game scenarios in gameplay and education. Through a taxonomy of diversity metrics and evaluation methods, we aim to bridge the current gaps in literature and practice, offering insights into effective strategies for measuring and integrating diversity in game scenarios. Our analysis highlights the necessity for a unified taxonomy to aid developers and researchers in crafting more engaging and varied game worlds. This survey not only charts a path for future research in diverse game scenarios but also serves as a handbook for industry practitioners seeking to leverage diversity as a key component of game design and development. 
    \end{abstract}
    %Employing a structured methodology, we delve into the essence of game scenario diversity. 

    \section{Introduction}
    \newcommand{\rev}[1]{\textcolor{red}{#1}}
    \newcommand{\unrev}[1]{\textcolor{pink}{#1}}
    
    \revised{\IEEEPARstart{V}{ideo} games have become a ubiquitous and influential media of entertainment, captivating millions of players and developers worldwide~\cite{Jovanovic2023Gamer}. Within the gaming world, one essential aspect that impacts player engagement and satisfaction is the diversity of game content~\cite{Yannakakis2018Artificial}. 
    As the domain of games continues to expand, there's a concerted move among researchers and designers towards devising effective methods to evaluate and amplify the richness of game content. This quest for diversity spans several domains, such as affective modeling~\cite{Yannakakis2008Entertainment,Yannakakis2023Affective}, procedural content generation (PCG)~\cite{Togelius2011SearchBased,Shaker2016Procedural,Summerville2018Procedural,Liu2021Deep,Guzdial2022Procedural}, game-based learning and therapy~\cite{Yannakakis2018Artificial,whyte2015designing}, and general machine learning~\cite{Risi2020Increasing}, each recognizing the unique contributions of varied game content to their respective fields.}
        
    \revised{For instance, in the domain of affective modeling, the pursuit of diverse game content is driven by the goal to capture a broader range of player emotions, thereby maximizing player satisfaction through tailored emotional engagements~\cite{Yannakakis2008Entertainment,Liapis2019Orchestrating,Yannakakis2023Affective}. Similarly, in the field of multi-agent systems, there's a strong emphasis on enriching the diversity of scenarios for researching and validating trustworthiness of autonomous agents~\cite{Sahin2020Catch,Liu2022unified,Sun2023Diversity,Guerrero-Romero2023Playing,hu2024games}.} Psychological research delves into diverse game content to craft gameplay experiences that are not only more immersive but also educational, aiming to enhance the pedagogical efficacy of educational games~\cite{Yannakakis2007Modeling,Wouters2011Measuring,Koh2011Computing,Park2019Generating}. 

    Among this multidisciplinary pursuit, PCG emerges as a pivotal innovation~\cite{Yannakakis2011ExperienceDriven,Preuss2014Searching,Shaker2016Procedural,Yannakakis2018Artificial,Summerville2018Procedural,DeKegel2020Procedural,Liu2021Deep,Guzdial2022Procedural,Yannakakis2023Affective}, revolutionizing game development by automating the creation of engaging content and thereby enhancing the player experience. The evolution of PCG reflects a shift from focusing solely on content quality to embracing a broader dimension of content diversity~\cite{Yannakakis2018Artificial,Gravina2019Procedural,Beukman2022Procedural}. Initially concentrated on generating high-quality content, such as playable levels and challenges, the field of game AI is now exploring new horizons where diversity is as crucial as quality itself~\cite{Preuss2014Searching,Shaker2016Procedural,Yannakakis2018Artificial,Gravina2019Procedural,Risi2020Increasing}.

    \revised{The importance of game diversity extends beyond academic discourse, resonating strongly within industry practices. Early attempts with PCG in pioneering games, such as \textit{Akalabeth} and \textit{Rogue}, laid the foundation for the Rogue-like genre, influencing countless successors, including mainstream titles like \textit{Diablo}, which introduced randomized dungeons and item drops~\cite{Shaker2016Procedural,Amato2017Procedural,Yannakakis2018Artificial}. The advent of commercial design tools ushered in a new era for PCG, shifting content creation from manual labor to automated processes. Modern classics like \textit{Elite Dangerous} and \textit{Minecraft} further exemplify this shift~\cite{Shaker2016Procedural,Yannakakis2018Artificial}, demonstrating the versatility and profound impact of PCG across various gaming genres and its critical role in delivering diverse and dynamic content~\cite{Amato2017Procedural,Dolfe2022MixedInitiative}.}
    \revised{Particularly, mixed-initiative approaches in PCG have gained increasing attention~\cite{liapis2016mixed,guzdial2022mixed}. These methods and tools not only assist designers in creating novel content through interactive visualizations but also emphasize the need for game diversity. By integrating human creativity with algorithmic generation, mixed-initiative PCG with diversity considerations ensures that the content produced is not only diverse but also rich in design possibilities, and provides insights to designers and players.}
    % \lan{ADD Mix-initiative}
    % This transition acknowledges the increasing player demand for novel and varied experiences within games, urging developers to weave diversity into the fabric of game content~\cite{Shaker2016Procedural, Gravina2019Procedural}. 
    
    \revised{As we dive into the definition of game content diversity and measurements, the lack of a systematic review of the metrics and evaluation methods used in academia and industry becomes evident, which motivated this survey. 
    Scenario diversity refers to the variety, richness, and variations in game scenarios. However, it can be defined across various scopes, such as inter-level and intra-level, and in different facets, including level, gameplay, rules, etc.~\cite{Liapis2019Orchestrating} 
    We also observe the \textit{multi-dimensional} nature of diversity measures\footnote{\revised{To simply the vocabulary in this survey, the phenomena of multiple diversity measures under one facet and multiple diversity measures across facets are both referred to as the \textit{multi-dimensional diversity measures}. The specific term \textit{multi-faceted diversity measures} is used to refer to the later case.}}: (i) under one game facet~\cite{Liapis2019Orchestrating}, various measures have been proposed for evaluating the diversity of different elements of the facet, sometimes also with different representations; (ii) different diversity measures across facets~\cite{Liapis2019Orchestrating} can be used together to offer a more comprehensive evaluation of content diversity from different perspectives, which can be referred to as the \textit{multi-faceted} nature of diversity measures. Researchers and developers often design their diversity measures to meet their specific needs. There's an absence of a taxonomy of existing diversity metrics and methods, and a standardized approach.}
    
    \revised{Because of the numerous game genres and multi-faceted nature of content~\cite{Shaker2016Procedural}, this survey focuses on \textit{scenario diversity}.
    The term ``\textit{scenario}" is specifically defined to include all interactive and non-interactive elements that contribute to the gameplay experience. This encompasses the game's scenes, map, level, mechanics, and narrative elements that players engage with directly or indirectly during play. We distinctly exclude non-digital aspects, such as traditional board games and audio-visual elements like sound and music. Our scope does not cover in-game entity characteristics, such as characters, races, behaviors, costumes, the game's overall difficulty level, terrain unless evaluated within a game context, and the scope of affective computing concerning game diversity. 
    Focusing on the diversity of these elements, our survey aims to provide a systematic review and a taxonomy of diversity measures, helping researchers and practitioners better understand the characteristics of diversity metrics and methods, and further assist them with the selection of appropriate scenario representations, diversity metrics and methods for their specific needs. 
    By discussing the multi-dimensional nature of diversity measures, our survey also seeks to bridge the gaps in the current understanding of diversity metrics and provide a robust foundation for future research.
    }
    %such as metric validation and comparison, and game development.}
    %By analyzing the existing diversity metrics and their applications, our survey also seeks to create a roadmap for future works. We also provide valuable insights into these metrics discussing the multi-facet of diversity measures. Additionally, we aim to guide researchers and developers in selecting the most appropriate representations, diversity metrics and methods for their specific game scenarios and game aspects. Ultimately, our survey aims to bridge the gaps in the current understanding of diversity metrics and provide a robust foundation for future research such as metric validation and comparison, and game development endeavors.

    \revised{The structure of the paper is as follows. Section \ref{sec:methodology} outlines the methodology employed in this survey. Section \ref{sec:taxonomy} presents our taxonomy.
    The representation of game scenarios is examined in Section \ref{sec:representations}. 
    Section \ref{sec:objective} summarizes the measures and approaches used for the objective evaluation on these representations.
    Section \ref{sec:subjective} presents subjective evaluation methods. The multi-dimensional nature of diversity is discussed in Section \ref{sec:dimension}. Section \ref{sec:discussion} discusses the implications and findings, and gives an outlook. Finally, Section \ref{sec:conclusion} concludes the paper.}
    
    \section{Methodology of Survey\label{sec:methodology}}
    
    %Though there exists a variety of content types~\cite{Shaker2016Procedural}, our survey focuses on the diversity of game scenarios.
    %In this research, the term ``\textit{scenario}" is specifically defined to include all interactive and non-interactive elements that contribute to the gameplay experience in video games. This encompasses the game's scenes, map, level, mechanics, and narrative elements that players engage with directly or indirectly during play. We distinctly exclude non-digital aspects, such as traditional board games and audio-visual elements like sound and music. Our scope does not cover in-game entity characteristics, such as characters, races, behaviors, costumes, the game's overall difficulty level, terrain unless evaluated within a game context, and the scope of affective computing concerning game diversity. 

    \subsection{Related work and novelty of the survey}
    For comprehensive coverage of terrain generation, readers are directed to detailed surveys by Voulgaris \textit{et al.}~\cite{Voulgaris2021Procedural}, Galin \textit{et al.}~\cite{Galin2019Review}, Raffe \textit{et al.}~\cite{Raffe2012survey}, and Valencia-Rosado \textit{et al.}~\cite{Valencia-Rosado2019Methods}. 
    %Similarly, while Kelly and McCabe's exploration of city generation features and techniques is acknowledged~\cite{Kelly2017Survey}, it is considered minimally relevant to the concept of game scenarios as defined in our survey and is thus not included. 
    \revised{Similarly, Kelly and McCabe's exploration of city generation features and techniques is acknowledged~\cite{Kelly2017Survey}.}
    Finally, discussions on dynamic difficulty adjustment are available in a survey by Mortazavi \textit{et al.}~\cite{Mortazavi2024Dynamic}.

    Complementing these resources, Kutzia and von Mammen survey procedurally generated buildings~\cite{Kutzias2023Recent}. Viana and Santos systematically reviewed dungeon generation methods~\cite{Viana2021Procedural}. However, to the best of our knowledge, none of these works mentioned evaluation metrics. Similarly, Gravina \textit{et al.} propose PCG through quality-diversity but rarely discuss the specific evaluation methods for it~\cite{Gravina2019Procedural}. A. Liapis~\cite{Liapis202010} reviews a decade's research trends in PCG, noting a clear increase in publications. Yet, the majority of these papers still lack thorough evaluation~\cite{Liapis202010}. Yannakakis and Melhart's systematic review~\cite{Yannakakis2023Affective} on affective computing in games intersects with our interest but from a distinct angle focused on affective computing and player modeling, which diverges from our core focus on diversity metrics in game scenarios. 
    In~\cite{Dolfe2022MixedInitiative}, a list of papers that used expressive range to evaluate PCG generators is summarized. However, these metrics only apply to expressive ranges. While various metrics have been proposed and used, a structured taxonomy for diversity evaluation remains a need, as echoed in the works by \cite{Shaker2016Procedural,Yannakakis2018Artificial,Yannakakis2011ExperienceDriven}.

    In this survey, our focus is drawn toward the evaluation of video game items and the multi-dimensional nature of diversity in game scenarios through the lens of diversity metrics. While the broader discussions of social diversity within video games, touching upon aspects such as cultural nuances and representation, have been extensively explored in existing literature~\cite{Sato2021CrossCultural, Passmore2017Racial, Harvey2019Becoming, InternationalGameDevelopersAssociation2022Developer}. For those interested in the diversity of character identity within video games, the work by To \textit{et al.}~\cite{To2018Character} is recommended. Our investigation delves into the diversity of game scenarios and components. This includes the variety and complexity of game elements, mechanics, and narrative elements that collectively shape the gameplay experience. By concentrating on the diversity of in-game elements and their assessment through various metrics~\cite{Gravina2019Procedural,Dolfe2022MixedInitiative}, we aim to contribute to the understanding of how diverse game scenarios can be quantitatively evaluated and categorized. This exploration does not directly address how diversity enhances player experiences but rather establishes a foundation for future research to link these quantitative assessments with qualitative player experiences, thereby enriching the dialogue around the role of diversity in video games.

    \subsection{Paper collection methodology}

\begin{figure}[htbp]
        \centering
        \includegraphics[width=.62\linewidth]{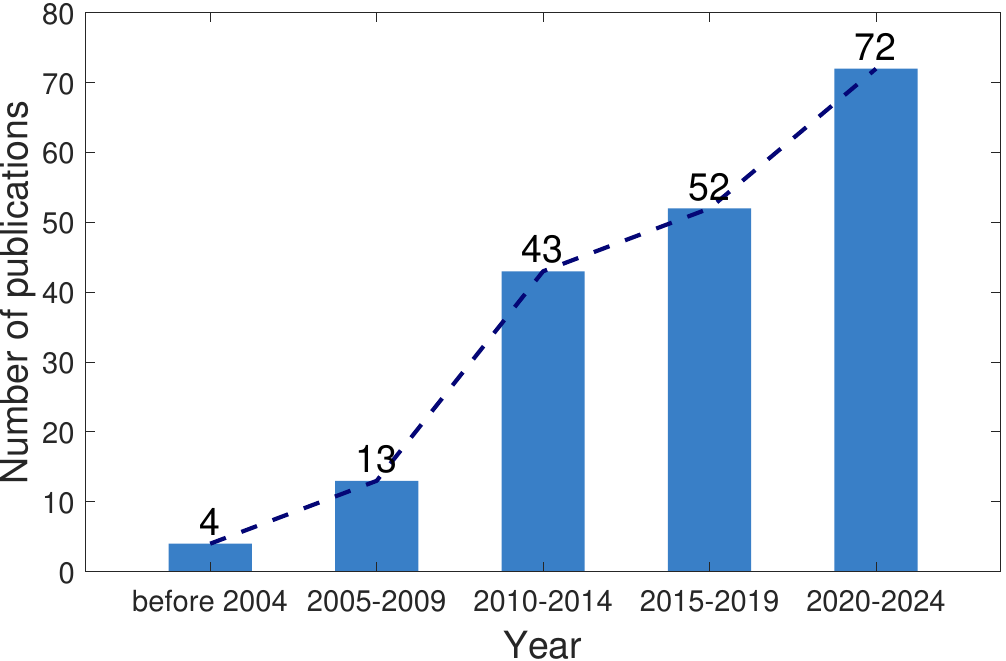}
        \caption{\revised{Number of papers organized by publication years.}}
        \label{fig:pub_year-label}
    \end{figure}

    \revised{We used the search term
    \textit{game AND $\sim$diversity OR novelty -bird -social diversity -gender -workplace -game theory -population diversity -dilemma -coalition -genetic diversity -deer -protein -game animals}. The presence of \textit{$\sim$diversity} allows for searching using synonyms of diversity. Then, we searched within the returned papers using the keywords \textit{diversity}, \textit{novelty}, \textit{similarity}, \textit{expressive}, \textit{divergence}, \textit{diverse}, and \textit{duplication}, according to \cite{Shaker2016Procedural} and \cite{Yannakakis2018Artificial}, to verify whether those papers truly involve game diversity-related content.} 
    %To ensure the comprehensiveness and reliability of our research, the literature review 
    \revised{The search was conducted using a range of reputable databases and sources, including \textit{Scopus}, \textit{ProQuest}, \textit{IEEE Xplore}, \textit{ACM Digital Library}, and \textit{Google Scholar}. Out of a total of 295 papers, 184 are within our scope and covered in this survey. As shown in Fig. \ref{fig:pub_year-label}, a growing research interest has been observed.}

    \section{Taxonomy\label{sec:taxonomy}}

    In the context of PCG and game design, this survey categorizes diversity evaluation into two primary categories -- \textit{objective evaluation} and \textit{subjective evaluation} -- offers a structured approach to assessing the diversity of game scenarios. This bifurcation not only simplifies the understanding of how diversity can be measured but also highlights the different lenses through which the effectiveness of game scenarios can be evaluated. Figure \ref{fig:taxo} illustrates our taxonomy.

        \begin{figure*}[hbp]
        \centering
        \resizebox{\linewidth}{!}{\begin{tikzpicture} 
    % \draw [help lines, color=gray!80] (-20, -10) grid (10, 2);
    \Large
    \node at (0, 0) {\textbf{Diversity Measure}};
    \draw[line width=4pt] (-6, -0.35) -- (4.8, -0.35);
    \draw[line width=4pt] (0, -0.35) -- (0, -6.8);
    
    \large
    \node at (-5.5, -1.4) {\textbf{Objective Evaluation (Sec. \ref{sec:objective})}};
    \node at (5, -1.7) {\textbf{Subjective Evaluation (Sec. \ref{sec:subjective})}};
    \draw[line width=3.2pt] (0, -1.7) -- (-9.8, -1.7);
    \draw[line width=3.2pt] (-5.5, -1.7) -- (-5.5, -7.3);
    \draw[line width=3.2pt] (0, -2) -- (8.3, -2);
    \draw[line width=3.2pt] (5, -2) -- (5, -6.6);

    \normalsize
    \node at (2.7, -4) {\textbf{Biological data (Sec. \ref{sec:bio})}};
    \node[text width=3cm] at (7.2, -3) {\textbf{Human evaluation (Sec. \ref{sec:human})}};
    \draw[line width=2.4pt] (5, -4.25) -- (0.35, -4.25);
    \draw[line width=2.4pt] (5, -3.5) -- (9.2, -3.5);

    \node at (-9.5, -2.6) {\textbf{Single-indicator (Sec. \ref{sec:single-ind})}};
    \node[right] at (-5.3, -3.1) {\textbf{Multi-indicator (Sec. \ref{sec:mult-ind})}};
    \draw[line width=2.4pt] (-5.5, -2.9) -- (-13.2, -2.9);
    \draw[line width=2.4pt] (-9.5, -2.9) -- (-9.5, -7.7);
    \draw[line width=2.4pt] (-5.5, -3.35) -- (-1, -3.35);
    \draw[line width=2.4pt] (-4.4, -3.35) -- (-4.4, -7.43);
    \draw[line width=1.6pt] (-4.4, -4) -- (-4, -4);
    \draw[line width=1.6pt] (-4.4, -5) -- (-4, -5);
    \draw[line width=1.6pt] (-4.4, -6) -- (-4, -6);
    \draw[line width=1.6pt] (-4.4, -7) -- (-4, -7);

    \small
    \node at (-12.2, -3.8) {\textbf{Comparison-based (Sec. \ref{sec:cmp})}};
    \node[text width=3.5cm] at (-7.5, -4.4) {\textbf{Non-comparison-based (Sec. \ref{sec:nocmp})}};
    
    \draw[line width=1.6pt] (-9.5, -4.05) -- (-14.6, -4.05);
    \draw[line width=1.6pt] (-9.5, -4.85) -- (-5.9, -4.85);

    \node[right, text width = 3cm] at (-4, -4) {\textbf{Expressive Range (Sec. \ref{sec:era})}};
    \node[right, text width = 3cm] at (-4, -5) {\textbf{Quality diversity (Sec. \ref{sec:qd})}};
    \node[right, text width = 2.8cm] at (-4, -6) {\textbf{Multi-objective (Sec. \ref{sec:mo})}};
    \node[right, text width=2.6cm] at (-4, -7) {\textbf{Other Analysis (Sec. \ref{sec:other})} };

    \footnotesize
    % comparison-based
    \node[right, text width=4.5cm] at (-14.6, -4.4) {$\circ$ Average Distance/Divergence};
    \node[right, text width=4.5cm] at (-14.6, -4.8) {$\circ$ Average Nearest Neighbor Distance};
    \node[right, text width=4.5cm] at (-14.6, -5.2) {$\circ$ Distribution of Similarity Values};
    \node[right, text width=4.5cm] at (-14.6, -5.6) {$\circ$ Dissimilarity Map};
    \node[right, text width=4.5cm] at (-14.6, -6) {$\circ$ Relative Diversity};
    \node[right, text width=4.5cm] at (-14.6, -6.4) {$\circ$ Novelty Score};
    \node[right, text width=4.5cm] at (-14.6, -6.8) {$\circ$ Computational Surprise};
    \node[right, text width=4.5cm] at (-14.6, -7.2) {$\circ$ Slacking A-clipped Function}; 

    % non-comparison-based
    \node[right, text width=4.5cm] at (-9.3, -5.2) {$\circ$ Standard Deviation};
    % \node[right, text width=4.5cm] at (-9.3, -5.6) {$\circ$ Coefficient of Variance};
    \node[right, text width=4.5cm] at (-9.3, -5.6) {$\circ$ \revised{Shannon's Entropy}};
    \node[right, text width=4.5cm] at (-9.3, -6) {$\circ$ Simpson Index};
    % \node[right, text width=4.5cm] at (-9.3, -6.8) {$\circ$ };
    % bio
    \node[right, text width=4.5cm] at (0.3, -4.6) {$\circ$ Facial/Head Expression};
    \node[right, text width=4.5cm] at (0.3, -5) {$\circ$ Heart Rate};
    \node[right, text width=4.5cm] at (0.3, -5.4) {$\circ$ Blood Volume Pulse};
    \node[right, text width=4.5cm] at (0.3, -5.8) {$\circ$ Skin Conductance};
    \node[right, text width=4.5cm] at (0.3, -6.2) {$\circ$ Borg Rating of Perceived Exertion};
    % human
    \node[right, text width=4cm] at (5.2, -3.8) {$\circ$ User Experience Questionnaire};
    \node[right, text width=4cm] at (5.2, -4.2) {$\circ$ Visual Comparison};
    \node[right, text width=4cm] at (5.2, -4.6) {$\circ$ Impression Curve};
    \node[right, text width=4cm] at (5.2, -5) {$\circ$ Interviews};

\end{tikzpicture}}
        %\resizebox{\linewidth}{!}{\input{figures/taxonomyGrant}}
        \caption{Taxonomy of measuring diversity of game scenarios.}
        \label{fig:taxo}
    \end{figure*}

    \textbf{Objective evaluation} in game scenarios is characterized by its reliance on quantifiable and directly measurable metrics without simulating gameplay. This category is further divided into \textit{single-} and \textit{multi-indicator} approaches, each offering unique insights into the diversity of content.
    
    \textit{Single-indicator approaches} focus on using a singular metric to evaluate diversity, within which, methods can be further classified into \textit{comparison-based} and \textit{non-comparison-based methods}. The metrics utilized by comparison-based methods are referred to as \textit{comparison-based metrics}, and focus on evaluating the diversity of game scenarios by comparing elements against one another. This comparison can illuminate the variance within a set of generated items, highlighting how each piece of content differs from the others. In contrast, metrics of non-comparison-based methods, \textit{non-comparison-based metrics} in short, analyze content independently of others, assessing diversity based on the presence, absence, or magnitude of specific features within each item.
    Single-indicator approaches, while straightforward, can sometimes offer a limited perspective on diversity due to their focus on a singular aspect of content.
    
    To capture a more holistic view of diversity, \textit{multi-indicator approaches} combine several metrics, including those used in single-indicator approaches. This category of approaches is enriched by the inclusion of \textit{visualization plots} and \textit{cumulative scores}, uniquely suited to the multi-indicator analysis including expressive range analysis and quality-diversity analysis. Visualization plots can illustrate the distribution and relationships of various features within the scenarios such as quality-diversity maps~\cite{Pugh2015Confronting} and expressive ranges~\cite{Smith2010Analyzing}, while cumulative scores aggregate multiple metrics into a single, comprehensive one such as quality-diversity score~\cite{Pugh2015Confronting} and coverage~\cite{Sarkar2021Generating}. \revised{Other analyses like t-distributed stochastic neighbor embedding (t-SNE)\cite{Maaten2008Visualizing} are also discussed as part of the multi-indicator approach.} 
    \def\toolong{This layered approach allows for a more nuanced understanding of scenario diversity, offering insights that might be missed when relying on a single indicator alone.}

    \textbf{Subjective evaluation}, in contrast to objective evaluation, involves \textit{human evaluation} and \textit{biological data} to assess the diversity of game scenarios. This method typically utilizes questionnaires, or biological data to gauge human responses to game content. Subjective evaluations capture the nuanced and sometimes intangible aspects of diversity that objective metrics might overlook, such as emotional impact, aesthetic appeal, and personal preference. However, subjective evaluation can be influenced by individual biases and is often more resource-intensive due to the need for participant involvement.
    
    Both objective and subjective evaluations play crucial roles in understanding the diversity of game scenarios. The former offers quantifiable and reproducible metrics, while the latter provides insights into human perception and experience. Together, these approaches offer a comprehensive review for assessing and enhancing the diversity of game scenarios, ensuring that game elements not only vary significantly from one another but also resonate with users on a personal level.
    \begin{revisedparas}

    In addition to the structured approach of objective and subjective evaluations as detailed previously, our taxonomy also embraces combined measuring methodology and strategy that contribute to a richer, multi-dimensional understanding of diversity. These methods are discussed in Section \ref{sec:dimension}, which seeks to combine multiple diversity metrics into a multi-dimensional analytical framework.
    \textbf{Multi-dimensional diversity evaluation} incorporates various analytical methods that span beyond traditional metrics, often leveraging the strengths of both objective and subjective evaluations. Examples include evaluating both content-centered and player-centered representations, and analyzing diversity both within individual levels and across different levels or scenarios.\end{revisedparas}

    Before delving into the specifics of various diversity evaluation approaches and metrics, Section \ref{sec:representations} reviews the representations of game scenarios upon which diversity is measured.

    \section{Representation of Game scenarios\label{sec:representations}}
    Game scenarios are originally represented as pixel-based images to users. Nonetheless, based on our comprehensive review, diversity assessments rarely leverage pixel-based representations directly~\cite{Giacomello2019Searching, Giacomello2018DOOM}. The representation of game components within game scenarios serves as a crucial foundation for measuring the diversity. Our survey captures the most widely-used representations in diversity measure and delves into two aspects: the definition of game components and their subsequent representation in game scenarios through \textit{content-centered} and \textit{player-centered} perspectives.
    %  TODO 晚点再看看这里如何精简

    \begin{figure*}[hbp]
        \centering
        \subfloat[2D levels]{\resizebox{\linewidth}{!}{\input{figures/mario-representations}}} \\
        \subfloat[2D maps]{\resizebox{0.345\linewidth}{!}{\input{figures/sokoban-representations}}}\hspace{5em}
        \subfloat[3D terrains]{\resizebox{0.3\linewidth}{!}{\input{figures/terrain-representations}}}
        \caption{Illustrative examples of game scenario  representations.}
        \label{fig:reprs}
    \end{figure*}
    
    %\subsection{Definition of game components}
    We focus specifically on empty and interactive components within games -- elements that offer substantial content, map, functionality, or purpose beyond mere audio elements that don't directly contribute to gameplay or narrative. These critical components include objects, levels, and game mechanics that players can interact with or that significantly influence the game's progress or experience. Notably, while actions such as character behaviors, character decisions, item selections, strategy formulations, and decision-making processes indeed contribute to the richness of \revised{player experience}, their diversification is categorized under a different facet of diversity, referred to as \textit{policy diversity} \cite{tan2023policy,Lupu2021Trajectory, Vinyals2019Grandmaster,Nam2022Generation, eysenbach2018diversity,florensa2017stochastic}. This aspect, which delves into the strategic and decision-making diversity within the gameplay \revised{in some specific game genre}, is further discussed in Section \ref{sec:genre}.
    %, distinguishing it from the direct measures of game components. 

    %\subsection{Content-centered vs. player-centered representations}\label{sec:contreps}
    % \def\toolong{With a clear understanding of the game components as defined above, this paragraph outlines two main types of commonly used representations to evaluate diversity in game scenarios: content-centered representation and player-centered representation. These representations are crucial for understanding and analyzing the dynamics and structure of games from both the content that comprises the game scenario and the interactions or behaviors of the players within that scenario.}

    \revised{We observe that the design of diversity metrics relies on the considered content type and its representation. Typically, objective contents within games such as levels and maps are considered as content-centered representations. These focus primarily on the structural elements of the game itself. On the other hand, studies that examine player interactions, such as strategies and the experiential aspect of gameplay fun, often employ player-centered representations. Some work (e.g., \cite{Wang2022Fun}) considered multiple representations to enrich the evaluation process. Fig. \ref{fig:reprs} illustrates different representations.}

    \subsection{Content-centered representation}
    Content-centered representation focuses on the structural aspects of the game. It includes various elements. \textit{Elements/patterns}~\cite{Sarkar2021Generating, Beaupre2018design} refer to the basic building blocks and the arrangement of those blocks within the game space, respectively. These can define the aesthetic and functional aspects of the game world. \textit{Strings}, which might represent textual or data-based elements within the game’s code or narrative structure.
    \textit{Top-down view}~\cite{Sfikas2022GeneralPurpose} is a kind of representation used in the $2$D or $3$D game scenarios, highlighting the geometry/shape of game space, especially. Some games apply the top-down view first and then utilize tiles to represent their levels/rooms~\cite{Park2019Generating}.
    \textit{Tiles}~\cite{Sarkar2021Generating, Biemer2021GramElites, Summerville2016Super, Summerville2016Learning,Summerville2018Expanding} indicate the spatial design elements of games, such as the construction of levels or maps through discrete units or ``tiles". \textit{Sequential cells}~\cite{Summerville2016Super,Summerville2016Learning}, suggesting a focus on the progression or order of tiles within a game, are often related to a more narrow-level design or structure in tiles.
    \textit{Height maps} can represent terrains~\cite{Frade2012Aesthetic,wulff-jensen2018deep,Raffe2012survey}. By recording the heights of a terrain at different coordinates, a complex terrain can be expressed by a $2$D array.
    \textit{Featured vectors} \cite{Preuss2014Searching} represent some specific game elements in a low-dimensional mathematical space.

    \subsection{Player-centered representation}
    Player-centered representation, on the other hand, concentrates on the player’s interaction with the game. 
    \textit{Behavior/action sequences}~\cite{Kondrak2005NGram} document the series of actions or decisions made by players during gameplay, providing insights into player strategies, preferences, and challenges encountered. Notably, these sequences are closely related to edit distance calculations, which measure the dissimilarity between them.
    \textit{Play traces}~\cite{Osborn2014Evaluating,Osborn2014GameIndependent} are records of player activities within the game, capturing the detailed paths players take, the choices they make, and the outcomes of those choices.

    Both representation types are essential for a comprehensive understanding of games, offering insights from the perspective of content creation and design, as well as from the player's interaction with and within the game environment.

    \section{Objective Evaluation}\label{sec:objective}
     Sections \ref{sec:metrics} and \ref{sec:objective approach} present metrics and approaches for objective evaluation\footnote{Due to page limit, the metrics and methods for objective evaluation are formulated using unified notations in \textit{Supplementary Material}.}, respectively.

    \subsection{Metrics for objective evaluation\label{sec:metrics}}

    \revised{Following our taxonomy, this section presents comparison- and non-comparison-based metrics for objective evaluation.}

    \subsubsection{Comparison-based metrics}\label{sec:com-metrics} 
    
    \revised{Comparison-based metrics are computed on representations of at least two game scenarios or through self-comparison using dissimilarity assessments including distance measures, divergence measures, and measures based on self-comparison. Table \ref{tab:com-obj} summarizes the metrics and the manipulated representations seen in literature, the corresponding references, and their pros and cons.
    }

\begin{figure*}[bp]
    \centering
    \begin{minipage}{0.35\linewidth}
    \vspace{0.04cm}
        \subfloat[Distance metrics for vectors]{\includegraphics[width=\linewidth]{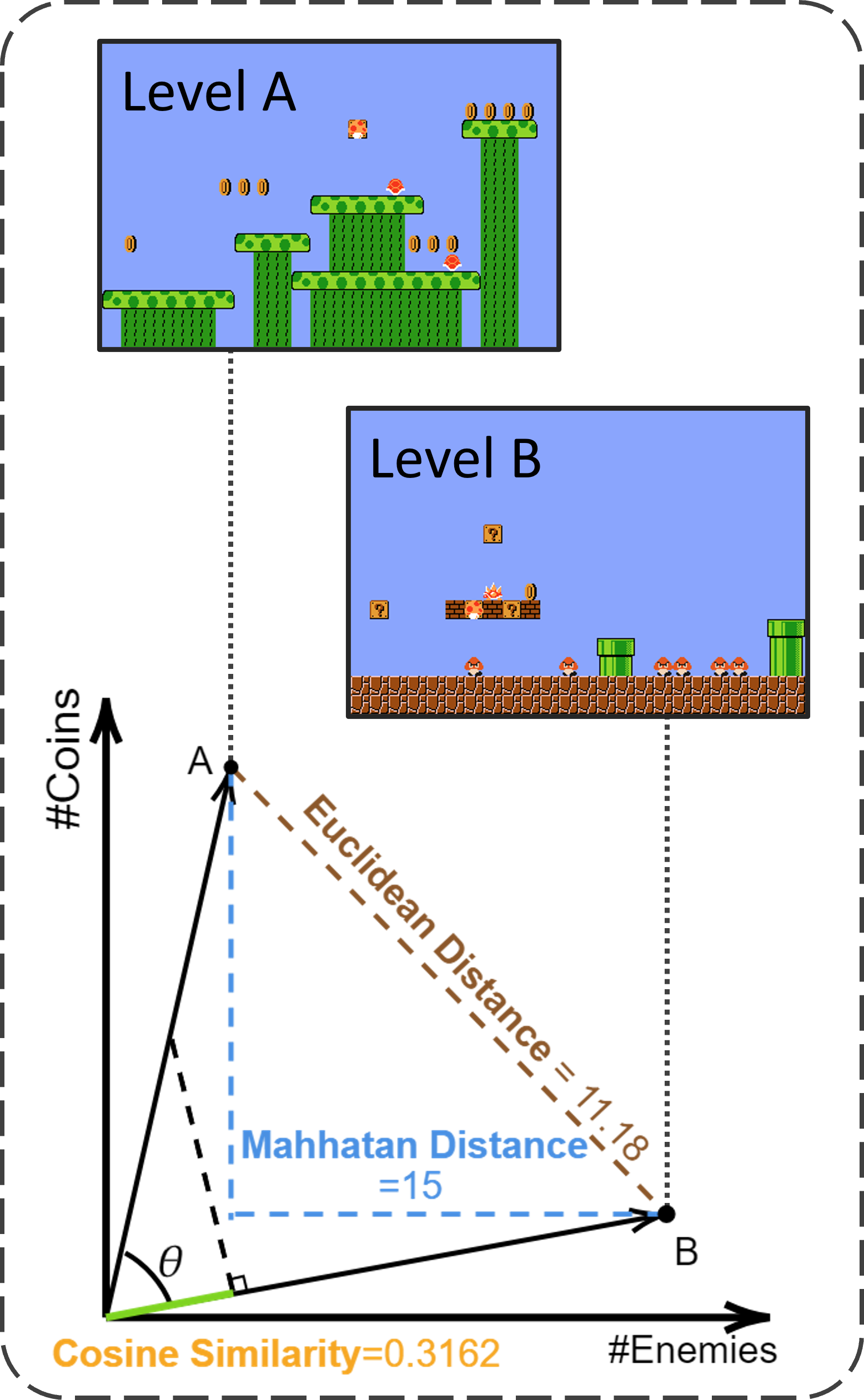}}        
    \end{minipage}
    \hfill
    \begin{minipage}{0.6\linewidth}
    \subfloat[Distance metrics for levels]{\includegraphics[width=\linewidth]{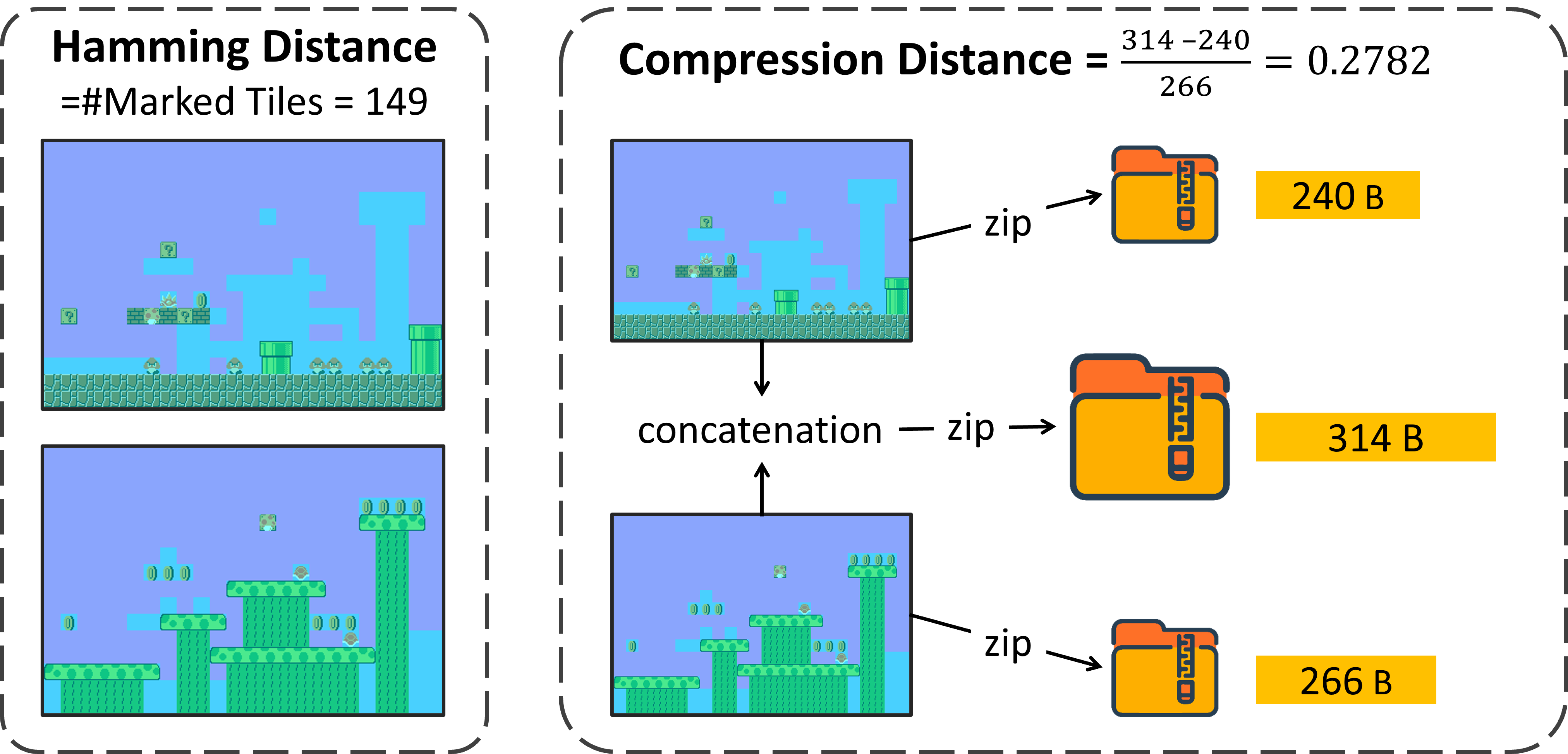}}
    \\
    \subfloat[Distance metrics for sequences. The image on the right-hand side is reused with permission from the authors \cite{wang2024fun}.]{\includegraphics[width=\linewidth]{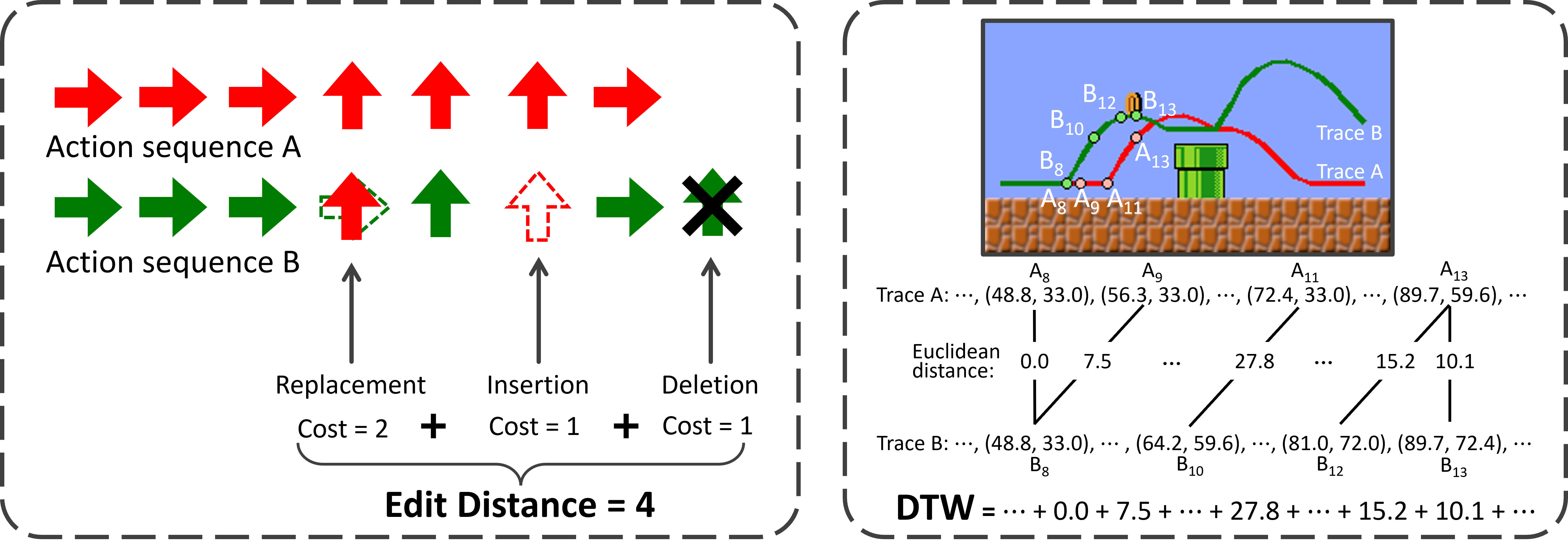}}
    \end{minipage}                
        \caption{Illustration of distance metrics, taking SMB levels as an example.}
        \label{fig:SMB-dis}
    \end{figure*}

    \paragraph{Distance measures} \textbf{Distance measures} are metrics used to quantify the dissimilarity between two game scenarios by calculating the distance between their representations. A number of distance-based metrics have been used to compare a pair of scenarios as they are simple yet straightforward. \revised{Fig. \ref{fig:SMB-dis} illustrates various distance-based metrics using \textit{Super Mario Bros.} (SMB) as an example.}

    \revised{\textbf{Cosine similarity/distance} is particularly effective when dealing with vector-based representations~\cite{Yuda2021Identification}.
    % \textbf{Manhattan distance}, also known as the \textbf{L1 distance},
    % is particularly applicable when comparing vector-based or sequence-based representations. 
    Oberberger \textit{et al.} apply \textbf{Manhattan distance} to measure the distance between potential fields of tactics for wargaming and real-time strategy games \cite{Oberberger2013Evolving}, and further use this metric in an evolutionary algorithm to search for a diverse set of tactics.
    \textbf{Euclidean distance}, also known as \textbf{L2}, is also widely applied in comparison-based diversity methods~\cite{Wang2023State,wulff-jensen2018deep,Lehman2011Abandoning,Lehman2013Effective,Nam2022Generation}.
    \textbf{Hamming distance} is commonly employed when dealing with discrete sequences of equal length, such as binary strings~\cite{Liapis2013Sentient,Preuss2014Searching,Wang2022Online,Wang2023State,Earle2022Illuminating,Zakaria2023Procedural,Wang2024Negatively,Liapis2015Constrained,Jiang2022Learning}. 
    It can also deal with $2$D or $3$D levels or maps by flattening the object into $1$D discrete sequences. \textbf{Compression distance} is a metric to measure the dissimilarity of two strings~\cite{Li2004Similarity}. It is based on a compression algorithm (e.g., gzip), 
    and is applied to compare 2D levels~\cite{Beukman2022Objective,Beukman2022Procedural}. 
    \textbf{Edit distance}, also known as \textbf{Levenshtein distance}, is a metric employed to quantify the dissimilarity between two strings or sequences~\cite{Levenshtein1966Binary} and offers insights into the minimal sequence of edits needed to align or transform one into the other. It is applied to measure the dissimilarity of action sequences~\cite{Beukman2022Objective,Beukman2022Procedural}, event sequences~\cite{Zook2012Automated} and flattened 3D maps~\cite{Awiszus2021WorldGAN, Dai2024Procedural}. \textbf{Constrained continuous edit distance} is a variation of the traditional edit distance that introduces constraints on the allowed edit operations~\cite{Chhieng2007Adaptive}. 
    Each operation (insertion, deletion, or substitution) is assigned specific constraints or costs, allowing for a more nuanced assessment of dissimilarity based on the application's requirements. Osborn and Mateas develop \textit{Gamalyzer}, a game-independent metric based on some refinements to constrained continuous edit distance to compare play traces~\cite{Osborn2014GameIndependent}.
    The subsequent research by Osborn \textit{et al.} confirm that \textit{Gamalyzer} aligns more closely with human perception of ``dissimilarity" and ``uniqueness", concepts akin to diversity, and claim that \textit{Gamalyzer} can help investigate ``\textit{Do players pursue diverse strategies?}"~\cite{Osborn2014Evaluating}.
    Unlike traditional distance measures, \textbf{Dynamic Time Warping (DTW)}~\cite{vintsyuk1968speech} measures the similarity between two sequences that may vary in wrapping policy, and allows for the elastic alignment of time series, accommodating temporal distortions. DTW is applied to compare play traces~\cite{Wang2022Fun}.
    \textbf{N-Gram similarity/distance} introduces a concept of measuring similarity and distance based on $n$-grams \cite{Kondrak2005NGram}. This approach demonstrates that traditional metrics like edit distance and the length of the longest common subsequence are special cases of $n$-gram distance and similarity, respectively. N-grams, which are contiguous sequences of $n$ items (e.g., characters or words).}

\begin{table*}[htbp]
    \centering
    \setlength{\tabcolsep}{3pt}
    \caption{\label{tab:com-obj}Comparison-based Metrics for Objective Evaluation of Scenario Diversity (Section \ref{sec:com-metrics})}
    \begin{tabular}{c|m{3cm}<{\centering}|m{2.7cm}<{\centering}|m{9cm}}
    \toprule
    \textbf{Category} & \textbf{Metric} & \textbf{Representations} & \textbf{Description, Pros, and Cons} \\ 
    \midrule
    \multirow{40}{*}{Distance} & Cosine similarity~\cite{Yuda2021Identification} & Featured vector & Measures the cosine of the angle between two non-zero vectors in a multi-dimensional space, focusing on orientation rather than magnitude. \newline
    \textbf{Pros:} Scale-invariant, works well with high-dimensional, sparse data. \newline
    \textbf{Cons:} Does not account for magnitude differences, less useful when magnitude is important. \\ \cline{2-4}
    
    & Manhattan distance~\cite{Oberberger2013Evolving} & Featured vector, height map, (continuous) sequence & Calculates the sum of absolute differences between corresponding elements of two vectors. \newline
    \textbf{Pros:} Robust to outliers, suitable for high-dimensional data, simple to compute. \newline
    \textbf{Cons:} Less effective when differences in feature scales are significant. \\ \cline{2-4}
    
    & Euclidean distance~\cite{Wang2023State, wulff-jensen2018deep, Lehman2011Abandoning, Lehman2013Effective,Nam2022Generation} & Featured vector, height map, (continuous) sequence  & Computes the straight-line distance between two points in multi-dimensional space. \newline
    \textbf{Pros:} Intuitive for evaluating magnitude of differences, captures geometric differences. \newline
    \textbf{Cons:} Sensitive to feature scales, requires normalized data, not robust to outliers. \\ \cline{2-4}
    
    & Hamming distance~\cite{Liapis2013Sentient,Preuss2014Searching,Wang2022Online,Wang2023State,Earle2022Illuminating,Zakaria2023Procedural,Wang2024Negatively,Liapis2015Constrained,Jiang2022Learning}.  & Tile, top-down view, string, sequences &  Counts the number of differing positions between two strings or binary vectors. \newline
    \textbf{Pros:} Efficient for categorical or binary data, easy to compute. \newline
    \textbf{Cons:} Limited to discrete data, ineffective for continuous features. \\ \cline{2-4}
    
    & Compression distance~\cite{Li2004Similarity,Beukman2022Objective,Beukman2022Procedural} & Subject to the compression algorithm &  Measures the distance based on the length of the compressed concatenation of two data objects compared to their individual compressions. \newline
    \textbf{Pros:} Captures both statistical and structural information, applicable across diverse data types. \newline
    \textbf{Cons:} Computationally intensive, dependent on compression algorithm. \\ \cline{2-4}
    
    & Edit distance~\cite{Levenshtein1966Binary,Beukman2022Objective,Beukman2022Procedural, Zook2012Automated, Awiszus2021WorldGAN,Dai2024Procedural} & Action/event sequence, flattened 3D maps&  Quantifies dissimilarity by counting the minimum operations (insertion, deletion, substitution) to transform one string into another. \newline
    \textbf{Pros:} Suitable for textual data and sequences, captures transformation costs. \newline
    \textbf{Cons:} Computationally expensive for long sequences. \\ \cline{2-4}
    
    & Constrained continuous edit distance~\cite{Chhieng2007Adaptive,Osborn2014GameIndependent,Osborn2014Evaluating} &Sequence, play traces &  Extends edit distance for continuous sequences, with constraints on alignment and transformation costs. \newline
    \textbf{Pros:} Suitable for time-series or continuous data, captures temporal or sequential differences. \newline
    \textbf{Cons:} More computationally expensive than standard edit distance. \\ \cline{2-4}
    
    & DTW~\cite{vintsyuk1968speech,Wang2022Fun} & Sequence, play traces &  Measures similarity between temporal sequences by optimally aligning sequences that may vary in timing or speed. \newline
    \textbf{Pros:} Effective for time-series data where events are not synchronized. \newline
    \textbf{Cons:} Computationally intensive, sensitive to noise. \\ \cline{2-4}
    
    & N-Gram similarity/distance~\cite{Kondrak2005NGram} & Sequence, n-grams & Compares sequences based on shared subsequences (n-grams). \newline
    \textbf{Pros:} Captures local patterns in sequences, less sensitive to global alignment. \newline
    \textbf{Cons:} Less effective for non-sequential data, limited to textual or sequential data. \\ 
    \midrule 
    
    \multirow{15}{*}{Divergence} 
    & Similarity~\cite{Alvarez2018Assessing,Alvarez2019Empowering,Alvarez2022Interactive} & Tiles, top-down view, vector, counts & Counts how many rows and columns are with different feature vectors. \newline \textbf{Pros:} Captures both vertical and horizontal features \newline \textbf{Cons:} Dependent on feature selection. \\ \cline{2-4}
    & KL divergence~\cite{Lucas2019Tile, Schubert2022TOADGAN,Awiszus2021WorldGAN, Dai2024Procedural,Csiszar1975I} & Patterns, tiles &  Measures how one probability distribution diverges from a second expected distribution. \newline
    \textbf{Pros:} Effective for probabilistic scenario comparisons. \newline
    \textbf{Cons:} Asymmetric, requires both distributions to have the same support. \\ \cline{2-4}
    
    & JS divergence~\cite{Wang2022Fun,dagan1997similarity} & Patterns, tiles &  A symmetric and smoothed version of KL divergence. \newline
    \textbf{Pros:} Symmetric, always finite, stable for comparing distributions. \newline
    \textbf{Cons:} More computationally expensive than KL divergence. \\ \cline{2-4}
    
    & Jaccard distance~\cite{Nam2022Generation,Murphy1996Finley} & Sets, feature sets &  Measures dissimilarity between sample sets as the complement of the Jaccard index. \newline
    \textbf{Pros:} Suitable for binary data or sets, intuitive for comparing sets or presence/absence of features. \newline
    \textbf{Cons:} Not useful for continuous data, ignores feature magnitude. \\ 
    \midrule 
    
    Self-comparison & Symmetry~\cite{Alvarez2018Assessing,Alvarez2019Empowering, Alvarez2022Interactive} & Tiles &  Evaluates the invariance of a scenario under specific transformations, indicating balance or symmetry. \newline
    \textbf{Pros:} Useful for assessing internal diversity and redundancy. \newline
    \textbf{Cons:} Less meaningful for asymmetric or dynamic scenarios. \\    
    \bottomrule
    \end{tabular}
\end{table*}

      \paragraph{Divergence measures} Sometimes, the distance between two game scenarios can be biased and is not able to correctly present the diversity of one or more game scenarios. Instead, comparing to a set of game scenarios is needed.
    \revised{\textbf{Similarity} quantifies the similarity between two contents by comparing the number of differing elements~\cite{Alvarez2018Assessing,Alvarez2019Empowering,Alvarez2022Interactive}.    
    Lucas \textit{et al.}~\cite{Lucas2019Tile} define the tile-pattern distribution and apply \textbf{Kullback-Leibler (KL) divergence}~\cite{Csiszar1975I} to compare the tile-pattern distributions of paired levels, computed as the frequency of each distinct tile pattern in a level/map.
    This metric is later widely applied in comparing SMB levels~\cite{Schubert2022TOADGAN, Dai2024Procedural} and Minecraft maps~\cite{Awiszus2021WorldGAN,Dai2024Procedural}.
    Wang \textit{et al.}~\cite{Wang2022Fun} apply \textbf{Jensen–Shannon (JS) divergence}~\cite{dagan1997similarity}, a symmetrized and smoothed variant of the KL divergence, to compare tile-pattern frequency distributions of SMB levels. 
    In the context of comparing sets of representations, \textbf{Jaccard distance}~\cite{Murphy1996Finley} provides a robust measure of distance based on the relative sizes of the intersection and union of the sets, and is applied to measure the number of different winning strategies in the work by Nam \textit{et al.}~\cite{Nam2022Generation}.}

    \paragraph{Measures based on self-comparison} A special case of comparison-based metric is comparing sub-sections of one single game scenario.
    \textbf{Symmetry}, as the name suggests, is a measure of how symmetrical a level/map is along both the horizontal and vertical axes~\cite{Alvarez2018Assessing,Alvarez2019Empowering, Alvarez2022Interactive}. Horizontally, it is computed by looking at pairs of rows starting at the center and moving outward and summing up the number of row positions that have the same tiles/content. Similarly, this is computed using pairs of columns for vertical symmetry. The final symmetry value for a level/map is the sum of the horizontal and vertical symmetry.

    \paragraph{Summary} 
    \revised{As summarized in Table \ref{tab:com-obj},} distance-based metrics such as Euclidean, Manhattan, and Hamming distances along with more complex measures such as cosine similarity and KL divergence, JS divergence, provide multiple choices for analyzing content diversity. The existence of this range of metrics showcases their ability to capture different aspects of diversity from simple positional differences to distributional diversity.
    However, the abundance of distance-based metrics also reflects underlying challenges. First, no such single metric universally captures all dimensions of diversity, so a multi-dimensional approach is necessary. Secondly, the choice of metrics is influenced by the data structure, and also significantly influences the perceived diversity, as each metric emphasizes different aspects of the game scenarios. For instance, while Euclidean distance might highlight spatial diversity, KL divergence could focus on the differences in the distribution of game elements. These different metrics, therefore underscore the complexity of measuring scenario diversity and point to an ultimate challenge: selecting the appropriate metrics that align with specific diversity dimensions of interest \revised{(cf. Section \ref{sec:dimension}).}

    For games with spatially rich scenarios, such as platformer games, metrics that characterize content diversity are essential. For instance, Manhattan distances can be applied effectively to measure the distance between game elements or levels. These distance-based metrics are particularly useful in vector representations. On the other hand, for behavior-specific diversity, edit distance or DTW provides a more detailed assessment of diversity by capturing the changes along with behavior sequences or play traces. Understanding how different distributions of game elements affect player experiences requires specific metrics, \revised{such as similarity, KL, JS divergence, or Jaccard distance. These tools help highlight how varied the gameplay can be, based on how game elements or levels are distributed. Additionally, for games aiming to offer a visually diverse experience, using symmetry as a metric can effectively measure how aesthetically varied a game scenario is.}

    \revised{In conclusion, while different comparison-based metrics enrich our objective evaluation of game scenario diversity, they also highlight the trade-offs between comprehensiveness, specificity, and practicality. The challenge appears not just in the selection of metrics but also in capturing the multi-dimensional nature of diversity within game scenarios.}

    \subsubsection{Non-comparison-based metrics}\label{sec:non-comparison-based metrics}
    In non-comparison-based methods, the emphasis shifts from measuring the dissimilarity between items to extracting meaningful features or characteristics from individual objects. 
    Non-comparison-based metrics are crucial in capturing relevant information for assessing diversity without relying on direct comparisons.
    The non-comparison-based metrics employed in assessing game-scenario diversity are summarized as follows.
    \revised{Table \ref{tab:noncom-obj} compares those metrics, highlights their corresponding representations, and outlines their pros and cons.}

    \begin{table*}[htbp]
    \centering
    \setlength{\tabcolsep}{3pt}
    \caption{\label{tab:noncom-obj}Non-comparison-based Metrics for Objective Evaluation of Scenario Diversity (Section \ref{sec:non-comparison-based metrics})}
    \begin{tabular}{m{5.3cm}<{\centering}|m{3cm}<{\centering}|m{9cm}}
    \toprule
    \textbf{Metric} & \textbf{Representations} & \textbf{Description, Pros, and Cons} \\ 
    \midrule
    Counting number~\cite{Summerville2016Super, Summerville2016Learning, Summerville2018Expanding, Nam2022Generation,Peng2021Detecting, Bonometti2020Theory,Melhart2022Arousal, Karth2017WaveFunctionCollapse, Beaupre2018design, Torrado2020Bootstrapping,Zakaria2023Procedural,Siper2023Controllable,Siper2022Path,Khalifa2016General,Frade2010Evolutiona,Szilas2014Objective,Szilas2007COMPUTATIONAL,Habonneau20123D,Partlan2018Exploratory} & Elements, patterns, events & Directly counts the number of elements, patterns, events, etc. \newline
    \textbf{Pros:} Works well for most games, useful when specific items are important. \newline
    \textbf{Cons:} Simple counting may not capture complexity or interactions between elements. \\ \cline{1-3}
    
    Linearity~\cite{Biemer2021GramElites, Smith2010Analyzing} & Level geometry & Measures how well the level geometry aligns with a straight line. \newline
    \textbf{Pros:} Suitable for platformer games, straightforward calculation. \newline
    \textbf{Cons:} Limited to games where linear structure is relevant. \\ \cline{1-3}
    
    Nonlinearity\cite{Sarkar2021Generating,Madkour2022NonTechnical} & Level geometry & Measures deviation from a straight line, opposite of the linearity metric. \newline
    \textbf{Pros:} Suitable for games with complex paths or exploration. \newline
    \textbf{Cons:} Not relevant for linear or fixed-path games. \\ \cline{1-3}
    
    Leniency~\cite{Smith2010Analyzing,Smith2010Tanagraa,Summerville2016Super,Summerville2016Learning,Summerville2018Expanding}& Tile-based levels & Describes how forgiving the level is to players, usually based on the number of helpful items. \newline
    \textbf{Pros:} Important for balancing difficulty, applicable to tile-based games. \newline
    \textbf{Cons:} Requires well-defined leniency metrics for non-tile-based games. \\ \cline{1-3}
    
    Density~\cite{Sarkar2021Generating,Biemer2021GramElites,Summerville2016Super,Summerville2016Learning,Summerville2018Expanding,Madkour2022NonTechnical} & Elements, patterns & Quantifies the frequency of specific items in a level. \newline
    \textbf{Pros:} Works for most games, useful when item frequency has significance. \newline
    \textbf{Cons:} May not account for item placement or interactions. \\ 
    \bottomrule
    \end{tabular}
\end{table*}

    In the analysis of the game scenario, the \textbf{counting number} metric signifies the total quantity of distinct components, objects, or entities within a specific context. Such a simple metric captures a wide array of items, such as the total number of jumps within a level and the subset of jumps deemed meaningful. A jump qualifies as meaningful if it is necessitated by the presence of an enemy or a gap, thus adding to the strategic complexity of the game~\cite{Summerville2016Super, Summerville2016Learning, Summerville2018Expanding}.
    The utility of the counting number metric extends across various domains, encompassing the evaluation of winning strategies~\cite{Nam2022Generation}, tracking alterations in knowledge graphs~\cite{Peng2021Detecting}, quantifying unique player actions~\cite{Bonometti2020Theory}, identifying distinctive features within datasets~\cite{Melhart2022Arousal}, and the enumeration of unique local patterns~\cite{Karth2017WaveFunctionCollapse}. Moreover, it is instrumental in analyzing the extremities of feature values, such as their maximums or minimums~\cite{Beaupre2018design}.
    Furthermore, several studies have employed the duplication percentage or the extent to which levels are replicated as a means to assess diversity across game levels, generators~\cite{Torrado2020Bootstrapping, Zakaria2023Procedural,Siper2023Controllable,Siper2022Path}, and within a broader dimension of games~\cite{Khalifa2016General}. 
    Regarding terrain generation in games, Frade \textit{et al.} group similar terrains and count the number of groups as a diversity measure~\cite{Frade2010Evolutiona}.
    They further introduce two metrics: the accessibility score based on the count of the inaccessible area and the obstacles edge length score~\cite{Frade2010Evolution, Frade2010Evolutiona}. Subsequently, they combine these metrics and develop a method that incorporates the weighted sum of these two metrics to evaluate terrain programs based on their height maps~\cite{Frade2012Automatic, Frade2012Aesthetic}.
    % The counting number metric is further utilized as a key evaluative evaluation method in numerous studies. 
    Szilas and Ilea \cite{Szilas2014Objective} delve into the assessment of diversity in interactive narratives, distinguishing between intra-diversity and global diversity. Intra-diversity refers to the count of different actions within a single playthrough, emphasizing session-specific diversity. Conversely, global diversity measures the array of unique actions across multiple playthroughs, thereby gauging the overall diversity encountered by players~\cite{Szilas2014Objective}. 
    This metric has also been utilized in \cite{Szilas2007COMPUTATIONAL,Habonneau20123D, Partlan2018Exploratory}.
    \textbf{Linearity} is a measure of how well the level geometry aligns with a straight line\cite{Smith2010Analyzing}. Each level is scored by summing the values of the distances from each point per unit to its expected value on the line, normalized by either the maximum linearity value or the maximum number of units~\cite{Biemer2021GramElites, Smith2010Analyzing}.
    In some cases, a unit may refer to the center point of each platform \cite{Smith2010Analyzing}, or every column in a level \cite{Biemer2021GramElites}.
    \textbf{Nonlinearity} is a measure based on the linear regression error when fitting a line to the structures within a segment
    % , employing a similar formulation to linearity 
    \cite{Sarkar2021Generating}.
    Notably, this metric is also applied to dungeon levels, where mission linearity captures the linearity of the mission structure and map linearity assesses the linearity of map layouts~\cite{Madkour2022NonTechnical}.
    \textbf{Leniency} is often calculated based on the quantity of specific items within a unit, such as enemies, gaps, or safety features, contributing to the game's difficulty or ease~\cite{Smith2010Analyzing, Smith2010Tanagraa}.
    Variations in calculating leniency might include the total number of enemies and gaps, offset by the number of rewards
    % , providing a nuanced view of the level's challenge
    ~\cite{Summerville2016Super, Summerville2016Learning, Summerville2018Expanding}.
    \textbf{Density} quantifies the frequency of specific items within a level, such as the number of tiles that are neither background nor path tiles~\cite{Sarkar2021Generating}, solid blocks~\cite{Biemer2021GramElites}.
    Density metrics in diversity evaluations can include the percentage of completable levels~\cite{Summerville2016Super, Summerville2016Learning, Summerville2018Expanding}, empty space~\cite{Summerville2016Super, Summerville2016Learning, Summerville2018Expanding}, reachable space~\cite{Summerville2016Super, Summerville2016Learning, Summerville2018Expanding}, interesting tiles among others~\cite{Summerville2016Super, Summerville2016Learning, Summerville2018Expanding}, and useless rooms~\cite{Madkour2022NonTechnical}.

    \paragraph{Summary} \revised{Non-comparison-based metrics offer a unique way to examine game scenarios by focusing on their individual features.
    %, such as the number of enemy types in a level or the complexity of story choices. 
    %They provide an overall picture of what each game scenario offers without direct comparison. 
    This approach raises an interesting point: even though these metrics start by examining scenarios individually, the insights they provide can lead to comparisons and their specific characteristics. For example, observing how strategies vary across levels or how narrative choices diversify gameplay. Essentially, those metrics also highlight characteristics of an individual level. When applied to a broader collection of levels, this metrics collection can better highlight a specific set designed to cater to player preferences. It is worth mentioning that multiple metrics can be used together, which adopts a comprehensive perspective on diversity by amalgamating various metrics. This methodology is enhanced through the integration of visualization plots and cumulative scores, elements particularly conducive to analyzing diversity through multiple indicators (cf. Section \ref{sec:mult-ind}).}

    \subsection{Objective evaluation approaches\label{sec:objective approach}}

    In the context of objective evaluation approaches, the distinction between single-indicator and multi-indicator approaches marks a pivotal methodology in assessing game scenario diversity (cf. Figure \ref{fig:taxo}). Single-indicator approaches focus on employing a singular metric to evaluate a specific aspect of diversity and offer a targeted and straightforward assessment method. In contrast, multi-indicator approaches adopt a more holistic view by integrating various metrics to provide a comprehensive analysis.
    %of diversity across multiple dimensions. 
    
    \subsubsection{Single-indicator approaches\label{sec:single-ind}}
    Within single-indicator approaches, we distinguish between comparison-based methods and non-comparison-based methods, each offering unique perspectives on evaluating game scenario diversity. Comparison-based methods concentrate on quantifying the dissimilarities between game elements or scenarios, leveraging mathematical metrics to measure the distance or difference between them. 
    On the other hand, non-comparison-based methods focus on the intrinsic characteristics of individual game components or scenarios without directly comparing them to others.
     
    \paragraph{Comparison-based methods\label{sec:cmp}}
    One fundamental approach for assessing diversity is through comparison-based methods, which calculate the distance or dissimilarity between contents and employ statistical techniques to access diversity measures. 
    %In the following sections, we refer to the metrics applied in comparison-based methods as \textit{comparison-based metrics}. 
    The outcome of these methods typically expresses a scalar value signifying the extent of diversity. 
    
    \textbf{Average distance/divergence} is a natural and widely-used metric to measure diversity~\cite{Zook2012Automated,Beukman2022Procedural,Earle2021Learning}.
    A notable portion of literature concerning tile-based levels or maps utilizes average Hamming distance to evaluate the scenario diversity. For example, Earle \textit{et al.} propose controllable PCG via reinforcement learning (RL) and computes average Hamming distance~\cite{Earle2021Learning}. Jiang \textit{et al.} extend this method to the 3D map generation of Minecraft, while the average Hamming distance is inherited as the diversity measure~\cite{Jiang2022Learning}.
    In works for online PCG on the SMB benchmark~\cite{Wang2022Online,Wang2023State,Wang2024Negatively}, average Hamming distance is applied as the scenario diversity measure.  
    Zook \textit{et al.} evolve tactical field care scenarios and evaluate the scenario diversity of their designed generator by computing edit distance between all scenarios in the evolution population, and illustrate the average and max distance over the by curves~\cite{Zook2012Automated}. 
    Mari\~{n}o \textit{et al.} suggest average compression distance is a proper measure of structural diversity \cite{Marino2015Empirical}. Later, Beukman \textit{et al.} also utilize average compression distance, along with average edit distance on the agent's action sequence, to evaluate the diversity of SMB levels~\cite{Beukman2022Procedural,Beukman2022Objective}. Awiszus \textit{et al.} also apply average edit distance to evaluate diversity, however, they directly compute the distance on the flattened string of 3D Minecraft maps~\cite{Awiszus2021WorldGAN}, supplemented by an average tile-pattern KL divergence.
    Nam \textit{et al.} represent role-playing game stages via event-parameter vectors, and employed average distance to evaluate the diversity of stages generated by RL policies~\cite{Nam2022Generation}. Two distance metrics are used in this work, to instantiate two ad-hoc diversity measures. The first is Euclidean distance which directly applies to the vector representation of stages. The second is the number of different winning strategies in two stages, which can be viewed as a variant of the Jaccard distance without normalization.
    To investigate the limitation of RL-based generators in online level generation, Wang \textit{et al.} evaluate the average Euclidean distance of \textit{latent vectors}~\cite{Wang2023State}.
    Wulff-jensen \textit{et al.} calculate mean square error and structured similarity index between randomly sampled pairs of generated 3D game maps, and report the average, median, standard deviation, and standard error of dissimilarity values~\cite{wulff-jensen2018deep}.

    Preuss \textit{et al.} define the objective-based diversity and visual-impression diversity as \textbf{average nearest neighbor distance} with different representations of levels~\cite{Preuss2014Searching}. The objective-based diversity evaluates several fitness functions to form a vector representation of game maps, while the visual impression diversity extracts visual featured vectors. Both applied Euclidean distance as the comparison-based metrics. 
    
    Coman and Mu\~{n}oz-Avila define \textbf{relative diversity} by addressing the average distance between a scenario and each scenario in a set, 
    to conduct case-based reasoning for playing real-time strategy games~\cite{Coman2010CaseBased, Coman2011Generating}.
    
    Instead of summarizing the similarity value into a scalar value, Khalifa \textit{et al.} show the \textbf{distribution of similarity values} via a curve plot of the probability density~\cite{Khalifa2017General}. 
    Another method with similar motivation, but different visualization is to plot the dissimilarity values with a heatmap, namely a \textbf{dissimilarity map}. 
    Marczak \textit{et al.} calculate a matrix of similarity values in gameplay audio of two players~\cite{Marczak2015Understanding}. This matrix is visualized through a heatmap, allowing for the visualization of both similar and dissimilar experiences between the players during gameplay. Through this approach, the matrix effectively highlights the degree of interaction and experience overlap or divergence between the participants.
    Schubert \textit{et al.} calculate the distances between generated levels and human-crafted levels and visualize the distances in a 2D heatmap~\cite{Schubert2022TOADGAN}. These works compare the dissimilarity between two sets of scenarios, rather than the dissimilarity within a set.
    
    \textbf{Novelty score} is usually used as a fitness function in evolutionary algorithms~\cite{Lehman2011Abandoning}. It measures the contribution to the population diversity of an individual. 
    % It is formulated as
    % \begin{equation}
    %   D(x|\mathcal{S}) = \frac{1}{k} \sum_{i=1}^k \delta(x, N_j(\mathcal{S})),
    % \end{equation}
    % where $N_j(\mathcal{S})$ is the $j$th nearest neighbor in $\mathcal{S}$ in terms of $\delta$.
    Similarly, Gravina \textit{et al.} define \textbf{computational surprise} as the behavioral difference between an item and its corresponding prediction \cite{Gravina2016Surprise}. 
    % It is formulated as
    % \begin{equation}
    %     D(x|\mathcal{S}) = \frac{1}{k} \sum_{j=0}^k \delta(x, M_j(\mathcal{S})),
    % \end{equation}
    % where $M$ is a model that predicts multiple possible behaviors based on the history, and $M_j(\mathcal{S})$ indicates the $j$-closest prediction to $x$, $k$ is a parameter determined empirically.
    Lehman and Stanley introduce \textit{novelty search} as an approach to discovering behavioral novelty \cite{Lehman2011Abandoning, Lehman2013Effective}, initially tested in maze environments.  Subsequently, Liapis \textit{et al.} apply novelty search with constraints in the field of PCG~\cite{Liapis2013Sentient, Liapis2013Enhancements}. The evaluation in novelty search relies on the novelty score, which calculates the average distance between an individual and its closest neighbor, as detailed in \cite{Liapis2013Sentient, Liapis2013Enhancements}. In their later research~\cite{Liapis2015Constrained}, they refer to Hamming distance as \textbf{visual diversity}, which involves comparing two maps on a tile-by-tile basis, presenting it as the comparison-based metric in novelty score. Later, Gravina \textit{et al.} introduce \textit{surprise search}, which utilizes a \textbf{surprise metric} based on Euclidean distance to measure the behavioral dissimilarity between an individual and its expected behavior \cite{Gravina2016Surprise}. Subsequently, they also introduce a weighted sum of the novelty score and surprise score, referred to as the \textbf{local competition score} compared with novelty score and surprise score \cite{Gravina2019Quality}. Sudhakaran \textit{et al.} propose MarioGPT to generate game levels from text prompts with a combination of large language model and novelty search, which employ the novelty score to evaluate the diversity \cite{Sudhakaran2023MarioGPT}.
    Novelty score is also applied with metrics such as tile-pattern KL divergence~\cite{Shu2021ExperienceDriven, Barthet2023OpenEnded}, Hamming distance~\cite{Melotti2019Evolving}, edit distance~\cite{Todd2023Level, Jackson2019Novelty} and compression distance~\cite{Beukman2022Objective}.

    \textbf{Slacking A-clipped function} is a method evaluating the diversity of content within a scenario\cite{Wang2022Fun}. It is devised for assessing online generated level segments, rewarding the moderate divergences of a new level segment regarding several previously generated ones. 
    % It is formulated as
    % \begin{equation}
    % \begin{aligned}
    % D(x_i|x_{i-1}:x_{i-n}) &=  \frac{1}{R} \sum_{k=1}^n \min\left\{1 - \frac{|g-\delta(x_i, x_{i-k})|}{g}, r_k \right\}, \\
    % r_k &= 1-\frac{k}{n+1}, \\
    % R &= \sum\nolimits_{k=1}^n r_k,
    % \end{aligned}
    % \end{equation}
    % where $n$ and $g$ are hyperparameters, representing the number of previous segments to be considered and the desired value of divergence, respectively. In this metric, segments are sequential while $x_i$ indicates the $i$th segment in the sequence.
    Wang \textit{et al.} embed both the tile-pattern JS divergence as a content comparison metric and the DTW as a gameplay comparison metric into this function to evaluate multi-dimensional diversity in SMB levels \cite{Wang2022Fun}.

    \paragraph{Non-comparison-based method\label{sec:nocmp}}
    
    Non-comparison-based methods often rely on metrics or counts related to specific characteristics or specific items within a game scenario. This could involve assessing the distribution of a particular item or attribute within the game scenario. The output of non-comparison-based methods is also usually a scalar value in single-indicator approaches, but these non-comparison-based metrics can also be combined in multi-indicator approaches (cf. Section \ref{sec:mult-ind}) with other formats of outputs.

    If the scenarios are represented as a single feature value, it is applicable to assess diversity via the \textbf{standard deviation}~\cite{Wang2021Keiki, Sorensen2016Breeding, Sombat2012Evaluating, Yannakakis2007Optimizing, Yannakakis2007Capturing}.
    % as formulated below~\cite{Wang2021Keiki, Sorensen2016Breeding, Sombat2012Evaluating, Yannakakis2007Optimizing, Yannakakis2007Capturing}:
    % \begin{equation}
    %     D(X) = \sqrt{\E\left[\Big( \phi(x) - \E[\phi(x)] \Big)^2\right]},
    % \end{equation}  
    % here, $\phi(x)$ is the feature value of a scenario $x$, and $\E$ signifies the expected value of the feature values across scenarios. 
    A vanilla standard deviation method is widely applied in multiple game scenarios with selected non-comparison-based metrics.
    Cook \textit{et al.} introduce analytical techniques for evaluating generator samples, emphasizing two essential aspects: the centroid, representing the average \textit{connectedness and density} score, and the standard deviation of the \textit{connectedness score} with a same initial solid chance parameter setting, which indicates the dispersion of the sample \cite{Cook2019General}. 
    Wang \textit{et al.} report the standard deviation bullet hell game barrages' (``Danmaku") \textit{shooting frequency}, \textit{mean momentum}, and \textit{(screen) coverage}~\cite{Wang2021Keiki}. They claim that possibly a larger standard deviation indicates better diversity. 
    
    Furthermore, Zhang \textit{et al.} design a coefficient of variance measure to evaluate the \textit{gameplay event diversity} while playing a game level, which is defined as the fraction of \textbf{standard deviation} and the mean value~\cite{Zhang2022Generating}. 
    Sorochan \textit{et al.} utilize four metrics for evaluating the different aspects of generated levels, which include \textit{gold total per level}, \textit{percentage collected per level}, \textit{total nodes explored}, and \textit{nodes per gold}. They also compute the standard deviation of each of these metrics across the generated levels. To visualize the results, they employ a boxplot for standard deviation and histograms to represent the distribution of interesting tiles and space within the game Lode Runner \cite{Sorochan2021Generating}.

    Another widely used method, behavior diversity, is originally proposed to measure ``diversity in opponent behavior over the games" based on the \revised{\textbf{standard deviation}} by Yannakakis \textit{et al.}~\cite{Yannakakis2005Generic,Yannakakis2007Optimizing}. 
    Later this method is also utilized in other kinds of games with distinct non-comparison-based metrics~\cite{Sombat2012Evaluating}. 
    In addition to behavioral diversity, spatial diversity is another crucial aspect explored by Yannakakis \textit{et al.}~\cite{Yannakakis2005Generic, Yannakakis2007Optimizing}. This method focuses on the configuration and composition of the game scenario. By segmenting a map into various feature values, such as nodes, tiles, cells, and other spatial elements, researchers can quantify the spatial diversity of a game scenario. The calculation of spatial diversity employs \revised{\textbf{Shannon entropy}}, a measure derived from information theory that provides a statistical means of evaluating the unpredictability or complexity of a system\cite{Togelius2010Towards, Ma2022AngleBased, Yannakakis2006Comparative, Yannakakis2005Generic, Sombat2012Evaluating}.
    Typically, the counting number is employed as the non-comparison-based metric in spatial diversity.

    In their works of 2005 and 2007~\cite{Yannakakis2005Generic, Yannakakis2007Optimizing, Yannakakis2007Capturing}, Yannakakis and Hallam introduce various quantifications of interest criteria, with a specific focus on two key diversity aspects: behavior diversity and spatial diversity. Behavior diversity is quantified by calculating the standard deviation of the \textit{time taken}, while spatial diversity is measured using the entropy of \textit{visited cells}. These metrics are applicable to a broad range of predator/prey computer games, as they are based on generic features common to this category of games. Later, the concept of spatial diversity measured by entropy of \textit{visited nodes} is also adopted in research involving ``fun" analysis by Yannakakis \textit{et al.}~\cite{Yannakakis2006Capturing, Yannakakis2006Comparative, Yannakakis2008Entertainmenta}. Pedersen \textit{et al.} compute the spatial diversity based on the gap count and visualize the count of each collected feature of player behaviors to evaluate the diversity of generated levels \cite{Pedersen2010Modeling}. Later, Sombat \textit{et al.} also use spatial diversity of \textit{maze features} and behavior diversity of the \textit{survival duration} to model interest in gameplay~\cite{Sombat2012Evaluating}. 
    
    \revised{
    In the 2010 Mario AI Championship\cite{Shaker20112010}, the concept of \textbf{spatial diversity}, measured by standard deviation, is adopted widely. They evaluate diversity on multiple levels based on \textit{gaps} and \textit{enemy placements}. Furthermore, they collect statistics on the \textit{numbers of coins}, \textit{rocks}, and \textit{powerups} as essential features to compare competitors in the championship, highlighting the importance of various gameplay elements in assessing diversity.    
    Togelius \textit{et al.} advocate that spatial diversity reflects the interestingness of StarCraft maps \cite{Togelius2010Towards}.
    Loiacono \textit{et al.} apply extracted frequency distribution from curvature profiles and speed profiles, to establish two diversity measures for the track of high-end racing games \textit{entropy} \cite{Loiacono2011Automatic}.
    Dutra \textit{et al.} define the diversity of a scenario as the entropy of event occurrences \cite{Dutra2022Procedural} and use entropy as both the reward to train RL-based level generators and the evaluation criterion.}

    \textbf{Simpson index} \cite{Simpson1949Measurement} is a classic measure of diversity by addressing the number of categories and element frequencies which is similar to spatial diversity. 
    %It ranged in $[0, 1)$. 
    The more uniform the frequency distribution is, the higher the Simpson index value.
    This measure is adopted by Berland \textit{et al.} to evaluate both game choice diversity and game outcome diversity \cite{Berland2023Joint}. 

            \begin{revisedparas}
       \paragraph{Summary} Single-indicator approaches provide foundational methods for assessing diversity within game scenarios by focusing on specific aspects of content and gameplay. 
       %These approaches are divided into comparison-based and non-comparison-based methods, each offering distinct insights into the compositional and experiential diversity of game environments.
    By utilizing these single-indicator approaches, researchers and game developers can gain a clear and quantifiable understanding of the diversity present within game scenarios. Although these methods concentrate on specific indicators, they significantly contribute to the broader objective of crafting engaging, dynamic, and diverse gaming experiences that captivate and challenge players across various dimensions. Combining multiple methods can lead to a more comprehensive evaluation of diversity, and it will be discussed in Section \ref{sec:dimension}.

    While single-indicator approaches provide valuable insights into specific aspects of game diversity, they also exhibit inherent limitations. Notably, they fail to capture the intricate interrelationships between different metrics and the synergistic effects that multiple elements may have on the overall gameplay experience. Consequently, these approaches might overlook more aspects of diversity that emerge from the interactions between various game components. To address these shortcomings and achieve a more holistic understanding of game scenario diversity, the next section explores multi-indicator approaches. These methods consider multiple dimensions of diversity simultaneously, offering a richer, more integrated analysis of game environments and player interactions.

    \end{revisedparas}

    \subsubsection{Multi-indicator approaches\label{sec:mult-ind}}

    %Another main branch of diversity evaluation methods is the multi-indicator approach. 
    A scenario is projected into a multi-dimensional feature space, where each dimension corresponds to a single diversity indicator.  The diversity is reflected by the distribution of the scenario in that feature space. We delve domain of multi-indicator approaches into expressive range analysis, quality-diversity analysis, and multi-objective analysis, each presenting a sophisticated lens through which game scenario diversity can be examined.
    The output of these methods can be scalar values indicating the degree of diversity or charts visualizing the diversity.

    % \rev{ERA and QD analysis often rely on a partition $\Pi$ of the feature (behavior) space, as defined by metrics in Section \ref{sec:metrics}. A partition of a space (set) $\F$ is a collection of subsets of $\F$ where the union of all items in $\Pi$ equals to $\F$, and any two items $\pi$ and $\pi'$ are non-overlapping, i.e., $\pi \cap \pi' = \varnothing$. In the context of multi-indicator approaches, the evaluation involves a set of samples $\S$ indexed by the partition $\Pi$. We denote the samples within $\S$ corresponding to a specific partition item $\pi$ as $\S(\pi)$, expressed as $\S(\pi) = \{x | x \in \S \land \f(x) \in \pi\}$. It's worth noting that the $\Pi$ can be a non-uniform partition.}
    
    \paragraph{Expressive range analysis (ERA)\label{sec:era}} 
    ERA is a widely adopted method for comprehending complex generative spaces by transforming intricate high-dimensional representations of artifacts into more accessible $2$D visualizations \cite{Smith2010Analyzing, Smith2008framework, Withington2023Right}. Table \ref{tab:ERA_metrics} summarizes the metrics and related works in ERA, while Table \ref{tab:ERA_visualization} summarizes the visualization techniques.

        \begin{table}[htbp] 
    \centering
    \caption{Metrics in ERA}
    \begin{tabular}{m{3.8cm}|m{4cm}}
    \toprule
    \textbf{Metrics} & \textbf{References} \\ 
    \midrule
    One/multiple objects (e.g., patterns and game elements) & \cite{Alvarez2022Interactive, Alvarez2018Assessing, Alvarez2019Empowering, Horn2014Comparative, Summerville2016Super, Summerville2016Learning, Summerville2018Expanding, Zakaria2023Procedural, Zakaria2023Start, Sarkar2021Generating} \\ \hline
    Color islands & \cite{Hald2020Procedural} \\ \hline
    Density of objects & \cite{Shaker2012Evolving, Summerville2016Super, Summerville2016Learning, Summerville2018Expanding, Alvarez2022Interactive, Alvarez2018Assessing, Alvarez2019Empowering, Horn2014Comparative, Zakaria2023Procedural, Jiang2022Learning, Sarkar2021Generating, Biemer2021GramElites} \\ \hline
    Linearity & \cite{Snodgrass2017Procedural, Biemer2021GramElites, Snodgrass2015Hierarchical, Snodgrass2016Controllable, Snodgrass2017Learning, Shaker2012Evolving, Alvarez2022Interactive, Alvarez2018Assessing, Alvarez2019Empowering, Dolfe2022MixedInitiative, Horn2014Comparative, Summerville2016Super, Summerville2016Learning, Summerville2018Expanding} \\ \hline
    Leniency & \cite{Snodgrass2017Procedural, Biemer2021GramElites, Snodgrass2015Hierarchical, Snodgrass2016Controllable, Snodgrass2017Learning, Shaker2012Evolving, Alvarez2022Interactive, Alvarez2018Assessing, Alvarez2019Empowering, Dolfe2022MixedInitiative, Horn2014Comparative, Summerville2016Super, Summerville2016Learning, Summerville2018Expanding} \\ \hline
    Graph complexity & \cite{VanderLinden2013Designing, Baghdadi2015procedural} \\ \hline
    Danger & \cite{VanderLinden2013Designing, Baghdadi2015procedural} \\ \hline
    Axiality & \cite{Baghdadi2015procedural} \\ \hline
    Generated paths & \cite{Baghdadi2015procedural, Zakaria2023Procedural, Jiang2022Learning, Zakaria2023Start} \\ \hline
    Shortest to winning paths comparison & \cite{Baghdadi2015procedural, Zakaria2023Procedural, Jiang2022Learning} \\ \hline
    Symmetry & \cite{Alvarez2022Interactive, Horn2014Comparative, Alvarez2018Assessing, Alvarez2019Empowering, Sarkar2021Generating} \\ \hline
    Similarity & \cite{Alvarez2022Interactive, Sarkar2021Generating, Alvarez2018Assessing, Alvarez2019Empowering} \\ \hline
    Entropy & \cite{Herve2021Comparing} \\ \hline
    Compression distance & \cite{Horn2014Comparative} \\ \hline
    Edit distance & \cite{Smith2012Expressive, Smith2011Launchpad} \\ 
    \bottomrule
    \end{tabular}
    \label{tab:ERA_metrics}
\end{table}

\begin{table}[htbp]
    \centering
    \caption{Visualization Techniques in ERA}
    \begin{tabular}{l|l}
    \toprule
    \textbf{Visualization} & \textbf{References} \\ 
    \midrule
    Histograms & \cite{Shaker2012Evolving, Sorochan2021Generating, Torrado2020Bootstrapping} \\ \hline
    Scatter plots & \cite{Madkour2022NonTechnical, Dahlskog2014Linear, Dahlskog2014Procedural, Snodgrass2021Studying, Sarkar2021Generating} \\ \hline
    Heatmap & \cite{Gravina2019Procedural, Dolfe2022MixedInitiative, Alvarez2022Interactive, Alvarez2018Assessing, Alvarez2019Empowering, Horn2014Comparative, Summerville2016Super, Summerville2016Learning, Summerville2018Expanding, Barthet2023OpenEnded} \\ \hline
    Density estimation & \cite{Snodgrass2021Studying, Summerville2018Expanding} \\ \hline
    Corner plots & \cite{Summerville2016Super, Summerville2016Learning, Summerville2018Expanding, Giacomello2019Searching} \\ \hline
    Density plots & \cite{Sfikas2022GeneralPurpose} \\ \hline
    Dissimilarity maps & \cite{Horn2014Comparative} \\ \hline
    Raincloud plots & \cite{Herve2021Comparing} \\ \hline
    Box plots & \cite{Horn2014Comparative} \\ \hline
    Codependency analysis & \cite{Cook2019General} \\ \bottomrule
    \end{tabular}
    \label{tab:ERA_visualization}
\end{table}

    ERA visualizes sets of levels in different scopes by employing two or more measurable metrics mentioned in Section \ref{sec:metrics}. Utilizing these metrics, each artifact is localized within a $2$D plot as expressive ranges, providing a concise overview of how the generated content is distributed in this metric-defined space~\cite{Withington2023Right}.
    The visualization can be implemented by a \textbf{scatter plot} showing each artifact's location in the metric-defined space, or by a \textbf{height map} with its grid corresponding to a partition of the feature space. The color of each cell in the height map represents the number or frequency of the scenarios in its corresponding subspace.
    By condensing the richness of generative spaces into visually interpretable representations, ERA facilitates a more intuitive understanding of the characteristics and variations.

    Building on Smith \textit{et al.}'s groundwork for evaluating game level\cite{Smith2008framework}, their later work introduces an \textbf{expressive range scatter plot}, focusing on \textit{linearity} and \textit{leniency} values for each level, thereby refining the analysis within specified ranges and deepening the understanding of the diversity in level set \cite{Smith2010Tanagra, Smith2010Tanagraa}. This also marks the emergence of a standardized method for ERA, which employs weighted hexagons to visualize how levels distribute across different parameters, offering insights into the diversity of game~\cite{Smith2010Analyzing}.
    Further advancements are made with the introduction of geometric similarity, utilizing edit distance on vectors encoding geometry types for each beat, providing a nuanced scale for assessing diversity~\cite{Smith2012Expressive, Smith2011Launchpad}.
    \revised{ERA underscores the ongoing interest in measuring the expressivity of a generator especially and, by extension, the diversity of generated levels, even though the term ``\textit{diversity}" is not explicitly used in their conventional sense.}
    The emphasis on ``\textit{expressivity}" as a means to evaluate generator diversity is further validated in the book~\cite{Shaker2016Procedural}, where the authors clarify that visualizing expressivity acts as a method to assess the diversity of the generator. This focus on expressivity and the subsequent development of evaluation methods illustrate a nuanced approach towards understanding and assessing the diversity of game levels.
    For a more straightforward comparison, Cook \textit{et al.} use expressive range to visualize changes in expressivity through the automatic optimization of generators, employing the tool \textit{Danesh} for this purpose~\cite{Cook2016automatic, Cook2022Danesh, Cook2016Danesh}. To select suitable metric pairs, Withington and Tokarchuk evaluate various metric pairs, ranking them to highlight the bias from improper metric selection~\cite{Withington2023Right}.

    Additionally, multiple video game genres are evaluated by ERA to measure diversity considering some genres-specific metrics. For room-based games, metrics like diameter, solidity, rooms, and skewness~\cite{Giacomello2019Searching, Giacomello2018DOOM}, as well as the feasibility score and a multi-indicator approach including plan compactness and average room compactness for top-down room generation, have been utilized~\cite{Sfikas2022GeneralPurpose}. In Minecraft, 12 metrics, including light, defense, functional metric, aesthetic metric, etc., are considered, with some showing a notably high correlation with human evaluation scores~\cite{Herve2021Comparing}. Moreover, the challenges of 3D level generation have been tackled through specific tasks like maximizing level diameter, ensuring door connections, and generating dungeon levels~\cite{Jiang2022Learning}.

    \begin{revisedparas}

    \paragraph{Quality-diversity (QD) analysis\label{sec:qd}} In the multi-indicator approach, QD algorithms play a pivotal role as a distinct category of evolution-inspired techniques. Their primary objective is to simultaneously optimize for both the quality and the diversity of solutions~\cite{Togelius2011SearchBased}.
    This dual optimization is achieved through \textit{behavior characterization}, which involves representing generated content using multiple metrics as detailed in Section \ref{sec:metrics}. \revised{Table \ref{tab:game_diversity_metrics} summarizes QD behavior characterization metrics.}
    The utilization of multiple metrics for behavior characterization inherently addresses multi-dimensional diversity, as each metric captures a different aspect or dimension of the content's behavior~\cite{Pugh2015Confronting, Pugh2016Quality}. By evaluating and promoting diversity across multiple dimensions, QD algorithms are able to explore a wide and varied feature space, capturing a diverse range of behaviors~\cite{Gravina2019Quality}. The concept of behavior characterization is thus crucial in driving the diversity component in QD analysis, ensuring that the solutions are not only high-quality but also diverse across multiple dimensions.
    \end{revisedparas}
    
\begin{figure}[htbp]
        \centering
        \includegraphics[width=.6\linewidth]{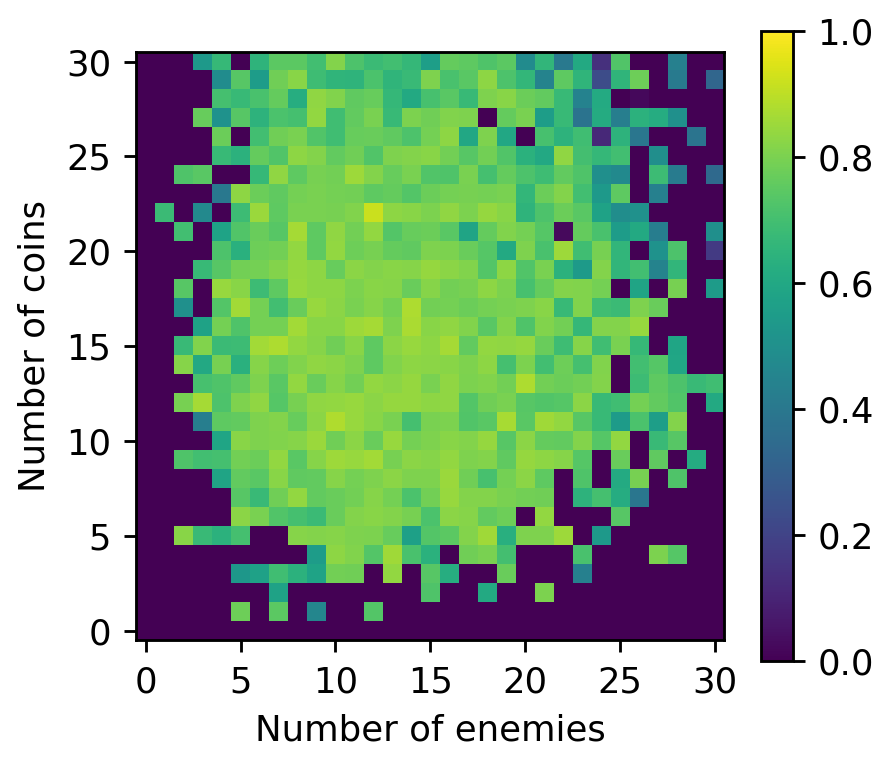}
        \caption{\revised{An example QD map made by mapping the elites of 5000 SMB levels generated by trained generative adversarial networks \cite{Volz2018Evolving}. The quality metric of this map is the slacking A-clip function \cite{Wang2022Fun}.}}
        \label{fig:qdmap}
    \end{figure}

    The \textbf{QD-score} serves as a conventional evaluation metric for QD algorithms. It is based on a partition of the feature space, aggregating the quality scores of the best content within each partitioned subspace~\cite{Pugh2015Confronting}. 
    The \textbf{coverage} measure calculates the proportion of subspaces occupied. This metric signifies the extent to which a specific algorithm such as MAP-Elites~\cite{Sarkar2021Generating}, successfully located solutions across the search space during the run.
    A \textbf{QD map}, often referred to as a QD-space or behavior space, is a graphical representation that visualizes the quality evaluation of a diverse range of solutions through a heatmap. The colors in the heatmap correspond to the quality evaluation values~\cite{Pugh2015Confronting}.  Figure \ref{fig:qdmap} shows an example QD map.

    In the early stages of QD research, evaluations typically rely on fitness values with \textit{tile entropy} and \textit{derivative-tile entropy} of the level, supplemented by the \textit{expressive ranges}\cite{Gravina2019Procedural, Khalifa2019Intentional}.
   This period marks the beginning of a broader exploration of game content and gameplay behaviors, encompassing diverse features within these domains. 
   \revised{For instance, the complexity of block and structural generation in Minecraft-liked platforms is examined~\cite{Soros2017Voxelbuild}, while bullet hell games utilize diversity metrics such as entropy, calculated from the first, second, and third derivatives of an agent’s action sequence, alongside risk, and distribution~\cite{Khalifa2018Talakat}. Subsequent studies extend this exploration to include game mechanics in games on general video game AI platform~\cite{Charity2020MechElites} and non-comparison-based metrics in platformer, puzzle, and dungeon games as shown in Table \ref{tab:game_diversity_metrics}.} For instance, in ``Baba is You'', a dynamic, rule-altering sokoban-style game, the rule activation distribution is depicted using distribution histograms~\cite{Charity2020Baba}. In deckbuilding, Fontaine \textit{et al.} introduce the QD algorithm to visualize the distributions of deck performance and density distributions of deck populations by average mana and mana variation~\cite{Fontaine2019Mapping}. Later, for the diverse strategies, the average hand size and the average number of turns, the max health differences are featured and visualized by a QD map~\cite{Fontaine2020Covariance, Zhang2022Deep}.

\begin{table}[htbp]
    \centering
    \caption{Metrics as Behavior Characteristics}
    \begin{tabular}{m{3.8cm}|m{4cm}}
    \toprule
    \textbf{Behavior Characteristics} & \textbf{References} \\ 
    \midrule
    Block/structural generation complexity & \cite{Soros2017Voxelbuild} \\ \hline
    Diversity metrics (entropy, risk, distribution KL divergence) & \cite{Khalifa2018Talakat,Fontaine2021Illuminating} \\ \hline
    Game mechanics in GVGAI & \cite{Charity2020MechElites} \\ \hline
    Density/frequency & \cite{Biemer2021GramElites, Sarkar2021Generating, Schrum2023Hybrid} \\ \hline
    (Non-)Linearity & \cite{Biemer2021GramElites, Sarkar2021Generating, Alvarez2019Empowering, Alvarez2018Assessing} \\ \hline
    Leniency & \cite{Biemer2021GramElites, Schrum2023Hybrid} \\ \hline
    Symmetry & \cite{Sarkar2021Generating, Earle2022Illuminating, Earle2021Learning} \\ \hline
    Similarity & \cite{Alvarez2019Empowering, Alvarez2018Assessing} \\ \hline
    Presence of game elements & \cite{Sarkar2021Generating, Alvarez2019Empowering, Alvarez2018Assessing, Fontaine2021Illuminating, Steckel2021Illuminating, Earle2022Illuminating, Earle2021Learning, Schrum2023Hybrid} \\ \hline
    Solution/path length & \cite{Steckel2021Illuminating, Earle2022Illuminating, Earle2021Learning} \\ \hline
    Presence of game actions & \cite{Fontaine2021Illuminating} \\ \hline
    Dynamics of actions (speed, time, size, jump entropy) & \cite{Warriar2019PlayMapper} \\ 
    \bottomrule
    \end{tabular}
    \label{tab:game_diversity_metrics}
\end{table}

% \begin{table}[htbp]
%     \centering
%     \textcolor{blue}{\caption{QD Map Visualizations and Metrics}
%     \begin{tabular}{m{3.8cm}|m{4cm}}
%     \toprule
%     \textbf{Visualization/Metric} & \textbf{References} \\ 
%     \midrule
%     QD maps (heat maps) & \cite{Fontaine2020Covariance, Sfikas2021Monte, Earle2022Illuminating, Fontaine2021Illuminating, Steckel2021Illuminating, Warriar2019PlayMapper, Fontaine2019Mapping, Zhang2022Deep} \\ \hline
%     Bar or histogram charts & \cite{Charity2020MechElites, Charity2020Baba, Khalifa2018Talakat} \\ \hline
%     Heat maps with density distributions & \cite{Fontaine2019Mapping} \\ \hline
%     QD score & \cite{Sarkar2021Generating, Earle2022Illuminating, Fontaine2021Illuminating, Sfikas2021Monte, Schrum2023Hybrid, Fontaine2020Covariance, Zhang2022Deep} \\ \hline
%     Coverage metrics & \cite{Sarkar2021Generating, Alvarez2018Assessing, Alvarez2019Empowering, Alvarez2022Interactive, Sfikas2022GeneralPurpose, Fontaine2021Illuminating, Sfikas2021Monte} \\ \hline
%     Percentage or number of filled cells/elites & \cite{Charity2020MechElites, Charity2020Baba, Earle2022Illuminating, Schrum2023Hybrid, Fontaine2020Covariance, Zhang2022Deep} \\ \hline
%     Evolution of QD-score or coverage (curve plots) & \cite{Sarkar2021Generating, Fontaine2020Covariance, Sfikas2021Monte, Fontaine2021Illuminating, Schrum2023Hybrid, Zhang2022Deep} \\ \hline
%     Frequency of runs found in each generator (heatmaps) & \cite{Biemer2021GramElites} \\ 
%     \bottomrule
%     \end{tabular}
%     \label{tab:QD_metrics}}
% \end{table}

    Typically, visualizations and cumulative assessments of diversity in QD research might encompass elements such as QD maps, which can feature heatmaps~\cite{Fontaine2020Covariance, Sfikas2021Monte, Earle2022Illuminating, Fontaine2021Illuminating, Steckel2021Illuminating, Warriar2019PlayMapper, Fontaine2019Mapping, Zhang2022Deep}, bar or histogram charts~\cite{Charity2020MechElites, Charity2020Baba, Khalifa2018Talakat} and heat maps with density distributions~\cite{Fontaine2019Mapping}. The QD score~\cite{Sarkar2021Generating, Earle2022Illuminating, Fontaine2021Illuminating, Sfikas2021Monte, Schrum2023Hybrid, Fontaine2020Covariance, Zhang2022Deep}, coverage metrics~\cite{Sarkar2021Generating, Alvarez2018Assessing, Alvarez2019Empowering, Alvarez2022Interactive, Sfikas2022GeneralPurpose, Fontaine2021Illuminating, Sfikas2021Monte}, and the percentage or number of filled cells/elites~\cite{Charity2020MechElites, Charity2020Baba, Earle2022Illuminating, Schrum2023Hybrid, Fontaine2020Covariance, Zhang2022Deep} also play a significant role. Most of these works also visualize the evolution process of QD-score or coverage via curve plots~\cite{Sarkar2021Generating,Fontaine2020Covariance,Sfikas2021Monte,Fontaine2021Illuminating,Schrum2023Hybrid,Zhang2022Deep}. Specifically, Biemer \textit{et al.}~\cite{Biemer2021GramElites} visualize the frequency of runs found bin for each generator through heatmaps, akin to ERA.
    
    In \cite{Gravina2019Procedural}, Gravina \textit{et al.} propose PCG through QD (PCG-QD) as a subset of search-based PCG. In their work, each component of PCG-QD is well-organized with discussion, and for more information beyond game generation about PCG-QD, one can refer to this work \cite{Gravina2019Procedural}.

    \paragraph{Multi-objective analysis\label{sec:mo}}
    Multi-objective optimization~\cite{Deb2008} is a technique to find a wide range of solutions approximating the Pareto front, which is the optimal set of solutions that are not dominated by any possible solution. 
    \def\toolong{The ``dominate" means a solution is not worse than the other in terms of any considered objective, while is better than the other in terms of at least one objective.} 
    \revised{There have been some works investigating multi-objective searches for game scenario generation, and diversity in terms of objective values is widely concerned \cite{Togelius2010Towards,Togelius2010Multiobjective,Togelius2013Controllable,Khalifa2020MultiObjective,zhang2024interpreting}.}
    Those works typically plot the locations of generated scenarios in the objective space by a scatter plot, to illustrate the diversity, we refer to this as the multi-objective scatter plot.
    In the context of StarCraft map generation, \revised{Togelius \textit{et al.} construct four computational metrics for the consideration of fairness, interestingness, aesthetics, and playability, then applies multi-objective optimization to generate maps with those characteristics \cite{Togelius2010Towards}.}
    Another work by Togelius \textit{et al.} emphasizes the importance of skill differentiation and interestingness \cite{Togelius2010Multiobjective}, both of which are closely tied to map diversity. They employ two fitness functions, choke points and path overlapping, to assess these aspects.
    In a later work \cite{Togelius2013Controllable}, they also employ a fitness function related to resource clustering, which is similarly modeled using Shannon's entropy. 
    Khalifa and Togelius conduct a comprehensive evaluation of their multi-objective level generator across various aspects \cite{Khalifa2020MultiObjective}. In the case of Binary, \revised{Zelda and Sokoban}, the evaluation involves calculating the number of regions and the improvement in path length within the generated map. \revised{Zhang \textit{et al.} generate Sokoban levels towards higher spatial diversity and emptiness through multi-objective evolution algorithm \cite{zhang2024interpreting}.}
    All of those works illustrate the diversity through the multi-objective scatter plot. 
    Wang \textit{et al.} train multiple generators with different trade-offs between fun score and diversity by varying a weight parameter \cite{Wang2024Negatively}. They also illustrate the diversity of those scenario generators by the multi-objective scatter plot showing their locations in the objective space. 
    \def\toolong{\textbf{Hypervolume}~\cite{zitzler1998multiobjective} is a frequently-used performance measurement in multi-objective optimization, which evaluates the overall performance of a solution set in terms of convergence and diversity among the considered objectives. Hypervolume computes the volume enclosed by the solution set in the objective space.}
    Besides the multi-objective scatter plot, the aforementioned work by Togelius \textit{et al.}~\cite{Togelius2010Multiobjective} also calculates \textbf{hypervolume}~\cite{zitzler1998multiobjective}, a frequently-used performance measurement in multi-objective optimization, which evaluates the overall performance of a solution set in terms of convergence and diversity among the considered objectives.  
    Ma \textit{et al.} incorporate the modeling approach in the work by Togelius \textit{et al.} \cite{Togelius2013Controllable} and propose a new multi-objective optimization algorithm to generate MegaGlest maps~\cite{Ma2022AngleBased}, supplemented by hypervolume. 

    % \begin{figure}
    %     \centering
    %     \includegraphics[width=\linewidth]{figures/PF2.pdf}
    %     \caption{Multi-objective scatter plot of non-dominated Sokoban levels generated by a multi-objective optimization algorithm. }
    %     \label{fig:enter-label}
    % \end{figure}
    
    % The generated map sets are also evaluated by the hypervolume. 
    \paragraph{Other analysis} \label{sec:other}
    
\begin{figure}[htbp]
        \centering
        \includegraphics[width=0.5\linewidth,trim=0 0 0 15,clip]{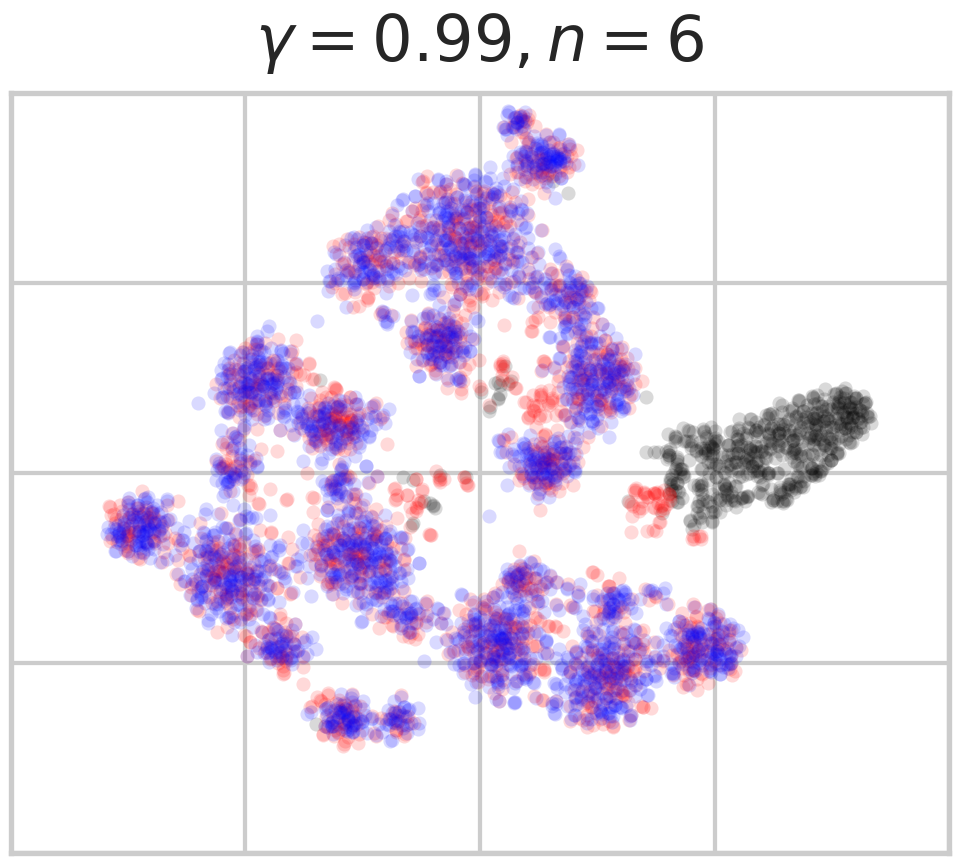}
        \caption{An example t-SNE embedding plot of generated levels, reused with the authors' permission \cite{Wang2023State}.}
        \label{fig:tSNE}
    \end{figure}

\begin{revisedparas}
    %As we have explored various methods and metrics for measuring diversity in game scenarios, 
    It has become evident that diversity in game scenarios cannot be encapsulated by a single-layered approach. 
    Instead, the complexity of game environments necessitates a multi-dimensional perspective to fully capture their diverse attributes. 
    \def\toolong{Building on the transition from single to multi-indicator analyses in our previous discussion, this part introduces a holistic visualization of overall diversity, which integrates multiple data layers, metrics, or behavior characteristics to present a more comprehensive understanding of the diversity within game scenarios. This shift towards a broader, more integrated analysis not only enriches our understanding but also enhances our ability to design and evaluate diverse gaming experiences.}
The \textbf{t-SNE} is a famous method to embed high-dimensional data into low-dimensional space, with the distance relationship approximately preserved~\cite{Maaten2008Visualizing}, and visualize the embedded points via scatter plots. Though this plot is not directly related to diversity, by plotting different sets of game scenarios in the same figure, their diversities can be compared.
\def\toolong{Snodgrass and Onta\~{n}\'{o}n train level generators with only $25$ training levels \cite{Snodgrass2017Procedural}. The generated levels are embedded with training levels using t-SNE and visualized. The plot shows that generated levels form a wide-spread distribution covering several training levels, indicating the generated levels are not overfitted but maintained diversity.
Similarly, Schubert \textit{et al.} generate levels with generative adversarial networks and visualize them along with training levels through the t-SNE plot \cite{Schubert2022TOADGAN}. }
For example, the work by Snodgrass and Onta\~{n}\'{o}n \cite{Snodgrass2017Procedural} and Schubert \textit{et al.} \cite{Schubert2022TOADGAN} both visualize generated levels along with training levels through t-SNE plots.
Wang \textit{et al.} plot different sets of level slices with the t-SNE plot (cf. Figure \ref{fig:tSNE}), and advocate the one with the most complex patterns may have the best diversity \cite{Wang2023State}. 
    Moreover, t-SNE is used to compare the ``exploration" of novelty search and random sampling \cite{Sudhakaran2023MarioGPT}. 

    \end{revisedparas}

    \begin{revisedparas}

        \paragraph{Summary} Multi-indicator approaches represent a sophisticated branch in the evaluation of game scenario diversity, moving beyond single-indicator methods to encompass a multi-dimensional analysis. These approaches project game scenarios into a multi-dimensional feature space, where each dimension corresponds to a different diversity indicator. The key techniques in multi-indicator analyses, such as ERA, QD analysis, multi-objective analysis, and other analyses, offer a nuanced perspective by visualizing and quantifying diversity through various sophisticated lenses.
        Overall, multi-indicator approaches allow for a more thorough and insightful analysis of diversity in game scenarios. By integrating multiple data layers and metrics, these methods offer a comprehensive view that captures the intricate interactions and variability within game environments. The transition from single to multi-indicator analyses marks a significant advancement in the field, paving the way for more detailed and meaningful evaluations of game scenario diversity.

            \end{revisedparas}
    \section{Subjective Evaluation \label{sec:subjective}}

    Subjective evaluation stands as a pivotal method for understanding the diversity within game scenarios, providing an in-depth look at how players perceive and interact with various elements of a game. This feedback is crucial for developers, illuminating both the strengths of the game and areas ripe for improvement. It ensures that the game not only offers a wide range of experiences but also resonates with a diverse audience. It fosters a deeper understanding of scenario diversity from the player's perspective, guiding future enhancements to better satisfy varied player needs. Typically, subjective evaluation measures a player's diverse experiences within a single game as an assessment of diversity in scenarios, but it can also extend to a horizontal comparison across different games within the same genre.
   
    \subsection{Human evaluation\label{sec:human}}

    Human evaluation plays a crucial role in assessing the diversity of game scenarios through qualitative assessments and annotations conducted by human evaluators.
    \revised{This method offers deep insights into the nuanced, subjective aspects of diversity that automated metrics might overlook. Common methods of human evaluation include various subjective assessments, and human emotions, and provide a comprehensive understanding of the qualitative dimensions of game diversity.}
    
    Using a \textbf{user experience questionnaire (UEQ)} to measure game scenario diversity during or after play provides critical insights into player perceptions and engagement. Questions targeting scenario variety, challenges, and interaction dynamics help assess the game's narrative and gameplay diversity. This method collects subjective feedback, offering a detailed view of the game's appeal across different player preferences. 
    \revised{The questionnaire can be enriched with human annotations like \textit{rank-based evaluation}, \textit{pairwise comparison}, and other evaluation methods, and it can be either a qualitative approach or a quantitative approach.}
    Osborn \textit{et al.} investigate the relation between their proposed ``dissimilarity metric" and humans' perception of dissimilarity, outliers, and uniqueness by a UEQ on a 7-Likert scale~\cite{Osborn2014Evaluating}. Similarly, Saurik \textit{et al.} implement a UEQ on a 7-Likert scale to estimate the novelty of a horror game~\cite{Saurik2023Evaluating}. UEQ is also used in \cite{Widiati2020User} with a similar process on a mobile-based game. 
    H\"am\"al\"ainen \textit{et al.} propose a framework that encompasses five facets for evaluating movement-based games \textit{after the interaction}~\cite{Hamalainen2015Utilizing}. The paper is structured around five central, gravity-related facets of user experience, identified based on their work and that of others: realism, affect, challenge, movement diversity, and sociality. Special attention is given to the aspect of movement diversity. 
    Partlan \textit{et al.} propose an evaluation method for interactive narratives \textit{during the interaction}, utilizing four key graphical representations: the scene graph, layout graph, script graph, and interaction maps \cite{Partlan2019Evaluation}. This evaluation method encompasses a playthrough of a scenario, an examination through a visual script editor, and a detailed walk-through of these representations. Each walk-through concludes with a set of questions aimed at scrutinizing the narrative structure, the scenario's interaction with the player, and other pertinent aspects. Notably, the second walk-through specifically focuses on visualizations provided by the script graphs and interaction maps, offering a deeper insight into the representation. The testers will naturally notice the aspects of diversity in game scenarios, and the designers can also lead them by asking diversity-related questions.   
    However, Szilas and Ilea \cite{Szilas2014Objective} suggest that an in-game questionnaire helps to better understand the player experience but may interrupt the overall experience.

    \textbf{Visual comparison} serves as an intuitive and immediate method for evaluating game scenario diversity, offering a direct way to observe differences by presenting a set of samples. This approach allows stakeholders, including developers, players, and researchers, to visually assess the variety and richness of game elements side by side. \revised{By examining these samples, one can easily identify the range of scenarios, interactive narratives, and gameplay mechanics, and highlight the diversity within a game or across different games through a qualitative approach and a quantitative approach.} Visual comparison not only makes the assessment of diversity straightforward but also provides concrete examples that can support discussions and decisions regarding game design and improvement. This technique is particularly effective in conveying the nuances of diversity that might be overlooked in textual or numerical evaluations, bringing a clear, impactful perspective on the visual and experiential variety present in game scenarios.
    In the work of Sarkar \textit{et al.}~\cite{Sarkar2021Generating}, examples of levels under different settings are presented to show diversity visually. \revised{Especially, they encode the absence or presence of the corresponding element as an $N$-digit binary number with the total $N$ of element types considered for the specific game through a quantitative approach. The generated levels with $N$-digits and other characteristics are displayed in \cite{Sarkar2021Generating} for readers to \textit{visually compare} the diversity of each level, which gives readers a chance to qualitatively evaluate the levels. Similarly, many other researchers use a qualitative approach to display their game diversity.} Steckel and Schrum also demonstrate the \textit{visual diversity} of their generated levels in their work \cite{Steckel2021Illuminating}. Sudhakaran \textit{et al.} plot the play traces of an agent on generated levels to illustrate the gameplay diversity \cite{Sudhakaran2023MarioGPT}. Holmg\r{a}rd \textit{et al.} present images of a level with different markers indicating different players' researched positions, to showcase the diversity of player decision-making styles \cite{Holmgard2016Evolving}. \revised{Dai \textit{et al.}, also apply visual comparison to evaluate their generated levels and maps\cite{Dai2024Procedural}.}
    
    The \textbf{impression curve}, introduced by K. Wejchert~\cite{wejchert1984elementy,Andrzejczak2020Impression}, is a concept that explores how space, time, and motion influence an observer's perception \revised{quantitatively}, particularly in the context of spatial images within different interior layouts, such as a street sequence. This method illustrates how various elements within a space impact an observer over time, without a specific measure but rather through a subjective scale from 1 to 10, where 1 represents a lack of architectural value and 10 signifies strong, meaningful architectural features.
    According to Andrzejczak \textit{et al.}~\cite{Andrzejczak2020Impression}, the impression curve has been applied in various studies to analyze and evaluate the diversity and significance of different spaces, including urban streets and rural landscapes, by mapping the observer's changing impressions over time.

    \revised{Conducting \textbf{interviews} provides a unique opportunity to gain in-depth insights into player experiences, perceptions, and preferences regarding game scenario diversity in a typical qualitative approach. Interviews facilitate direct conversations with players}, allowing for a comprehensive exploration of the nuanced ways in which different game elements are received and interpreted by individuals. Interviews can uncover detailed feedback on specific scenarios, reveal player strategies, and highlight emotional responses that other evaluative methods might not capture. 
    Schrum \textit{et al.} conduct a user study to investigate the experience of using their proposed interactive game design tool \cite{Schrum2020Interactive}. Despite the diversity is not a mandatory question, they receive some comments regarding diversity.
    
    \subsection{Biological data\label{sec:bio}}
    
    \def\toolong{Measuring the diversity of a video game using biological data presents a novel approach to understanding the complexity and variety within game scenarios, and narratives.}
    \revised{By analyzing biological data, such as heart-rate changes, researchers can quantify the diversity of interactions in a game, provide insights into the diversity of the gaming experience, and offer a unique perspective on how well the game simulates varied real-world scenarios or accommodates different player strategies. This could involve examining the variety of available strategies, or the complexity of the game's world. %Such an analysis can provide insights into the diversity of the gaming experience, and offer a unique perspective on how well the game simulates varied real-world scenarios or accommodates different player strategies.
    }

    \textbf{Facial} or \textbf{head expressions} are critical non-verbal communication cues that convey a wide range of emotions, thoughts, and intentions without the use of words. Facial expressions result from one or more motions or positions of the muscles beneath the skin of the face. 
    Shaker \textit{et al.} employ a range of features to assess player experiences, including \textit{game level (content) features}, \textit{gameplay behavioral features}, and \textit{head movement features} \cite{Shaker2013Fusing}. Analyzing players' \textit{head expressivity} in response to specific in-game events provides insights into the diversity of gameplay experiences through a mixed quantitative and qualitative approach. To evaluate these experiences, they administer four alternative forced-choice questionnaires to players after playing games with different feature values. These questionnaires allow players to rate their preferences for engagement, challenge, and frustration states. The use of \textit{pairwise preferences} in this assessment minimized the influence of factors such as personality and culture on self-reported player experiences, as these emotions are often closely linked to diverse gameplay experiences.
    \revised{Elor \textit{et al.} visualize players' performance, physiological responses, neural responses, and facial movements during the foundation and challenge protocols through a quantitative approach~\cite{Elor2022Gaming}, complemented by an emotion survey, which qualitatively highlights that the game sessions could evoke a diverse range of emotions.}
    
    Yannakakis \textit{et al.} have conducted a series of studies aimed at modeling entertainment~\cite{Yannakakis2007Modeling,Yannakakis2006Comparative,Yannakakis2008Entertainment,Yannakakis2006Capturing, Yannakakis2006Modeling, Yannakakis2008Entertainmenta}, by analyzing children's physiological responses during play in physical interactive playgrounds. These studies focus on capturing and modeling children's affective states \revised{through a quantitative approach}, particularly through measurements of \textbf{heart rate} (HR), \textbf{blood volume pulse} (BVP), and \textbf{skin conductance} (SC), during physical gameplay. Yannakakis \textit{et al.}'s research, which utilizes these physiological signals to model entertainment values, provides a foundational basis for exploring the relationship between game diversity, physiological responses and ``fun" experiences~\cite{Yannakakis2007Modeling,Yannakakis2006Comparative,Yannakakis2008Entertainment,Yannakakis2006Capturing, Yannakakis2006Modeling, Yannakakis2008Entertainmenta}. By selecting game benchmark levels based on diversity measures, one aligns with Malone’s principles of intrinsic qualitative factors—challenge, curiosity, and fantasy--that are essential for engaging gameplay. \revised{The variability, statistics and dynamic changes in HR, BVP, and SC during diverse gameplay scenarios can offer insights into how different game elements trigger varying levels of engagement and emotional arousal quantitatively.} For instance, games that introduce unexpected challenges or novel scenarios might elicit stronger physiological responses, indicating heightened curiosity or excitement, while those offering rich, fantasy-driven contexts could engage players on deeper emotional levels. Therefore, by analyzing the correlations between physiological responses and the scenario diversity, researchers can better understand how variations in gameplay contribute to the overall entertainment experience, and offer a quantitative measure to supplement traditional qualitative assessments. 
    %of game diversity. 

    \revised{\textbf{HR} and the \textbf{borg rating of perceived exertion (RPE)} are pivotal metrics for evaluating the intensity and physical exertion of players in virtual reality (VR) games that involve physical exercise during their gameplay quantitatively. Yoo \textit{et al.} conduct a study where they track players' percentage of maximum HR alongside the RPE for multiple VR games~\cite{Yoo2017Evaluating}, presenting their findings in a plot that also incorporates questionnaire responses through a qualitative approach.}

    \section{Multi-dimensional Diversity Evaluation}\label{sec:dimension}    
    \def\toolong{In the dynamic world of games, the pursuit of diversity in game scenarios presents a multi-dimensional challenge that touches upon a comprehensive and detailed approach, transiting from simple numerical evaluations to more complex and subjective approaches. }

    The pursuit of diversity in game scenarios presents a multi-dimensional challenge, transiting from numerical evaluations to more complex and subjective approaches. 

     \subsection{Which representation to use?}

    \revised{The selection of appropriate representations for measuring diversity is directly influenced by the \textit{facets} under consideration. There exists a clear mapping between facets and their corresponding representations, which plays a crucial role in the accurate assessment of game scenario diversity. For facets related to player interaction, referred to as player-centered representation in Section \ref{sec:representations}}, such as play traces~\cite{Osborn2014GameIndependent,Osborn2014Evaluating,Wang2022Fun}, action/behavior sequences~\cite{Beukman2022Procedural,Beukman2022Objective}, the chosen representations are inherently tied to the temporal and behavioral aspects of gameplay. Conversely, content-related facets are represented through content-centered representations such as tiles~\cite{Sarkar2021Generating, Biemer2021GramElites, Summerville2016Super, Summerville2016Learning,Summerville2018Expanding}, top-down views~\cite{Sfikas2022GeneralPurpose}, featured vectors~\cite{Preuss2014Searching}, and content sequences~\cite{Sorochan2021Generating}, reflecting the static or structural elements of the game environment.

    The choice of metrics for evaluating diversity is significantly informed by these representations. For instance, DTW is a comparison-based metric closely associated with temporal sequences, making it particularly suited for analyzing player behavior, where the timing and order of actions carry meaningful information~\cite{Wang2022Fun}. On the other hand, metrics for content-related diversity, such as tile-based KL divergence~\cite{Lucas2019Tile}, and Hamming distance~\cite{Preuss2014Searching,Liapis2013Sentient}, focus on comparing static features or configurations within game levels. These content-based metrics excel in quantifying the variation in level design, content features, or the presence of specific game elements.
    
    This approach to selecting representations and corresponding metrics also underscores the multi-dimensional nature of diversity in game scenarios. By tailoring the analytical framework to the specific facets of diversity, researchers can derive more meaningful and actionable insights into how diversity manifests within both the player experience and the game content. As the field progresses, the exploration of new representations and metrics suitable for untapped facets of diversity, such as music or narrative structures, promises to further enrich our understanding of what makes game environments engaging and diverse. 
    %This evolution in diversity measurement methodology highlights the importance of a comprehensive and flexible approach, capable of adapting to the diverse and dynamic nature of games.
    
    \subsection{Which evaluation approach to choose?}

     Diversity metrics used in objective evaluation can be categorized into \textit{intra-diversity}, \textit{inter-diversity}, and \textit{overall diversity}.
     \def\toolong{In objective evaluation approaches, especially, diverse methodologies are employed and diversity can be categorized into three \revised{categories}: \textit{intra-diversity}, \textit{inter-diversity}, and \textit{overall diversity}. 
   This structured framework reflects the various perceptions of diversity among researchers who held different research interests.  }

    % \textcolor{blue}{In the nuanced exploration of game scenario diversity, the methodology extends beyond mere numerical scoring to encompass a richer, more layered understanding. This approach reflects the inherent complexity and subjectivity in perceiving diversity within game environments.} Among the varied strategies for measuring game scenario diversity, a distinction can be drawn using concepts such as \textbf{intra-diversity}, \textbf{inter-diversity}, and \textbf{overall diversity}.

    \subsubsection{Intra-diversity}
    It focuses on the diversity present within individual components, levels, maps, or sessions of a game~\cite{Wang2022Fun,Zook2012Automated}. This category of measures illuminates the range of game elements, or options available in a singular context, scenario, level, or map, akin to assessing the variety of actions within a single period of gameplay or the unique elements within a specific level. The goal is to capture the richness and depth offered in isolated segments of gameplay, providing insight into the moment-to-moment variety that players might encounter.

    \subsubsection{Inter-diversity}
    Inter-diversity, on the other hand, examines the diversity across different components or sessions~\cite{Preuss2014Searching}. This perspective considers the variance between multiple playthroughs, levels, or game modes, offering a broader view of the diversity present in the game as a whole. 

    \subsubsection{Overall diversity}
    Overall diversity is employed as a holistic value or an overall visualization that captures the entirety of a game's diversity. This metric and/or diagram aim to quantify the game's total diversity, merging both the intra- and inter-diversity perspectives into a comprehensive evaluation. It serves as a summary indicator of the game's ability to offer players varied and engaging experiences, reflecting the combined effect of all diversity aspects.
    
    By incorporating these diverse metrics, intra-diversity, inter-diversity, and overall diversity, researchers can provide a multi-dimensional analysis of game scenarios. This methodology acknowledges the layered and subjective nature of diversity, aligning more closely with the intricate ways in which players experience and appreciate variety in gaming. The distinction between the scopes of diversity, coupled with an overall evaluation, highlights the complexity of game design and the importance of a systematic approach to measuring diversity. For instance, Zook \textit{et al.} utilize the edit distance to evaluate a single scenario and then compute the average and maximum value representing both intra- and inter-diversity for their generated scenarios, supplemented by a plot of diversity value over iterations for the overall diversity\cite{Zook2012Automated}.

    %\subsection{A structured methodology of diversity evaluation}
    %\subsection{Selecting the suitable metric}
    \subsection{Which methodology to choose?}
    % \subsection{Which metrics to choose?}

    This section discusses in a logical sequence that mirrors an analytical progression of diversity evaluation, illustrating appropriate metric pair dimensions toward different interests.
    
    \revised{\subsubsection{Multi-facet diversity metrics} Evaluating multi-faceted diversity in games involves introducing a range of metrics that consider various facets of the gaming experience. This comprehensive approach includes examining both player behavior and game content to assess scenario diversity. } Yannakakis \textit{et al.} have led the way in assessing diversity through measures like behavioral diversity and spatial diversity~\cite{Yannakakis2005Generic, Yannakakis2007Modeling, Yannakakis2006Capturing, Yannakakis2007Capturing}. These metrics, focusing on the variations in opponent behaviors and the game content~\cite{Yannakakis2005Generic, Yannakakis2007Modeling, Yannakakis2006Capturing, Yannakakis2007Capturing}, highlight the richness of diversity within individual game elements. This methodical approach provides insights into the layers of game content, emphasizing the significance of each game element's contribution to the overall gaming experience.

    \subsubsection{Multi-dimensional diversity metrics} 
    
    Building on the foundation established by facet-specific metrics, the methodology extends to the multi-dimensional implementation of metrics.
    \revised{Unlike the previously discussed multi-faceted metrics, this approach pairs metrics implemented in varied relationships, such as the combination of measuring diversity \textit{within individual levels} and \textit{between different levels.}}
    This multi-dimensional methodology includes a metric implemented in levels of a generator and in levels of multiple generators~\cite{Beukman2022Procedural, Horn2014Comparative}, a metric implemented in generated levels and in-between generated levels and training/target levels~\cite{Todd2023Level, Siper2022Path, Siper2023Controllable}, multiple metrics implemented in pairs of level and in addressing the expressivity of a generator\cite{Jiang2022Learning}, or multiple metrics considering diversity within generator and across multiple generators~\cite{Wang2024Negatively}.
    % \revised{Through the applications of these multi-dimensional approaches, these works illuminate the different approaches such as the combination of inter- and intra-diversity to comprehensively understanding diversity within specific dimensions. }
    \revised{These examples show a comprehensive and multi-dimensional approach that deepens different understandings of game diversity in different scopes.}

    \subsubsection{Visualization of diversity metrics} Finally, the synthesis and portrayal of game scenario diversity through the visualization of metrics mark the culmination of this structured methodology. The use of expressive ranges and curves for comparative analysis, either against training data benchmarks or within generator outputs in the context of PCG, is instrumental. These visual representations are vital in articulating the extent and depth of diversity across various systems, seamlessly connecting theoretical concepts with their practical applications in game design.
    
    Undoubtedly, measuring game diversity is a multi-dimensional endeavor. While it remains unclear which method can precisely capture the essence of  diversity, there is a hopeful anticipation for diverse perspectives to illuminate the concept of diversity from various angles. This aspiration towards a multi-dimensional understanding encourages continuous exploration and dialogue within the field, fostering a richer comprehension of how diversity shapes the gaming experience.

    \section{Discussion and Outlook\label{sec:discussion}}
    This section starts with the quest for the optimal degree of diversity within game scenarios, and then discusses the critical need for robust validation methods for diversity evaluation metrics, advocating for empirical studies that correlate these metrics with actual player experiences and preferences. The interplay between policy diversity and scenario diversity is examined, highlighting how AI strategies and diverse game scenarios influence the overall gaming experience and AI trustworthiness.
    Moreover, the influence of game genre on diversity evaluation is discussed, emphasizing the unique challenges and opportunities presented by different genres in achieving diverse game content. Finally, we identify and discuss gaps in current research and practice, including the disparity between academic theories and industry application, the potential of AI in automating diversity evaluation, the need for interdisciplinary approaches, and educational evaluation methods. 
    This section not only illuminates the challenges at hand but also reveals the opportunities for bridging gaps between theoretical frameworks and practical applications.

    \subsection{What is the desired diversity degree?}
    The examination of game scenario diversity in the literature often goes beyond mere quantification. Many studies prioritize the evaluation of dissimilarity between generated and predefined target game levels, sometimes addressing both dissimilarity and diversity. Todd \textit{et al.} highlight the importance of balancing similarity and dissimilarity within generated and target sets, suggesting that a moderate level of dissimilarity can be beneficial~\cite{Todd2023Level}. Some other research focuses on closely aligning generated levels with human-created examples or training datasets, (e.g., \cite{Liu2019Automatic,Kumaran2019Generating,Park2019Generating}).

    Zakaria \textit{et al.} examine specific in-game objects, such as the levels' \textit{entropy}, \textit{empty tile percentage}, and \textit{idle crates frequency}, to analyze how dissimilar procedurally generated levels are from those designed by humans~\cite{Zakaria2023Procedural}. Similarly, Torrado \textit{et al.} explore differences between generated and human-designed games by visualizing the count of levels alongside average Hamming distance and the diversity of game tiles~\cite{Torrado2020Bootstrapping}. 
    Wang \textit{et al.} propose a metric addressing intra-diversity of each level at a moderate value, engaging a fun experience \cite{Wang2022Fun}.
    Siper \textit{et al.} introduce another perspective on diversity~\cite{Siper2023Controllable}, distinguishing between inter-diversity, which measures the dissimilarity between generated levels and goal-set levels, and intra-diversity, which considers the uniqueness within each generated map. This differentiation is instrumental in understanding the complexity of diversity in game design and reflects the range of approaches within the field.

    % The literature showcases multitudes of perspectives and methodologies aimed at understanding and measuring diversity in game scenarios. 
    These varied approaches each offer unique contributions towards the collective goal of creating game scenarios that are engaging, diverse, and well-balanced. Through the lens of these studies, the field navigates the complex dynamics between generated content and predefined benchmarks, exploring both dissimilarity and alignment with human-designed ones.

 \subsection{Validation of diversity evaluation}

    Exploring the correlation between diversity metrics and their relevance across various gaming contexts has become a crucial area of research, aiming to bridge the divide between academic findings and practical applications within the game development industry. This area of research not only assesses the impact of varied game scenarios on player enjoyment and satisfaction~\cite{Yannakakis2007Capturing, Yannakakis2008Entertainment, Yannakakis2006Comparative, Huang2011Evaluating, InternationalGameDevelopersAssociation2022Developer, Nogueira2012Modeling} but also emphasizes the influence of these metrics on enhancing player experiences. Insights derived from such analyses are critical in creating immersive and engaging gaming environments that meet or exceed player expectations.

   % The practical application of diversity metrics within the gaming industry, especially with the advent of AI and machine learning, prompts a critical examination of how these metrics can inform game design decisions. Although collaborative efforts among researchers, game developers, and the gaming community are pivotal for refining these metrics, discussions on the correlation between specific metrics are sparse.

    Further complicating this field is the observation that traditional metrics like the mean squared error and the structural similarity index can yield contrasting results when comparing content generated by different algorithms~\cite{wulff-jensen2018deep}. This discrepancy underscores the challenge of selecting appropriate metrics and suggests that a singular metric may be insufficient for a comprehensive analysis of game level diversity.

    Despite the growing body of work exploring the relationship between diversity metrics and player perception, the direct analysis of this interplay remains relatively unexplored. Most existing user studies have concentrated on the capacity of these metrics to evoke player emotions, rather than on their ability to foster diverse content or experiences. Notably, the research by Osborn \textit{et al.} confirms that \textit{Gamalyzer} aligns more closely with human perception of ``dissimilarity" and ``uniqueness", concepts akin to diversity~\cite{Osborn2014Evaluating}. Fontaine \textit{et al.}~\cite{Fontaine2021Illuminating} conduct a study assessing the perceived similarity between generated levels and human-designed levels, aiming to determine if the KL divergence can model perceived similarity effectively. Other research extends similar investigation into the context of player emotions~\cite{Marino2015Empirical, McAllister2013Improving, Sombat2012Evaluating, Yannakakis2007Capturing}, underlining the significance of aligning diversity metrics with player perceptions to create more engaging and satisfying gaming experiences. This evolving discourse highlights the need for a comprehensive approach that considers both the objective and subjective dimensions of game diversity, emphasizing the critical role of player feedback and advanced analytics in shaping the future of game design.

\subsection{Game genre driven diversity evaluation }\label{sec:genre}
    %Diverse Agent Behavior }
    
     \revised{One notable aspect pertains to the evaluation of diversity within different game genres. A number of games or game-based platforms are available for researching AI and PCG~\cite{Hu2023Gamebased}. While certain game genres have been extensively researched, others have received relatively limited attention. This discrepancy underscores the need to explore diversity evaluation methods tailored to specific genres, as what constitutes diversity may vary significantly between genres. For example, many games, especially in \textit{real-time strategy games}, \textit{fighting games}, and \textit{role-playing games}, involve the interaction between players and agent-controlled characters. Though the diversity of agent behaviors is beyond our scope of ``scenario diversity", sometimes it can crucially affect the game experience. Therefore, we briefly discuss some measures regarding the diversity of agent behaviors as follows.}

    There is a wide range of multi-agent RL research proposing different diversity measures for agent behaviors. 
    For example, Vinyals \textit{et al.} plot the distribution of the average number of each unit built by Protoss agents throughout league training, normalized by the most common unit, to evaluate the agent~\cite{Vinyals2019Grandmaster}.
    Lupu \textit{et al.} define the diversity of agent behaviors as the average JS divergence with respect to the decision-trajectory distribution, from the individual policies to the average policy  \cite{Lupu2021Trajectory}.
    \revised{Some works measure the diversity of agent behavior in a multi-objective sense. 
    Zheng \textit{et al.} visualize game-playing policies through scatter plots with winning rate and exploration score \cite{Zheng2019Wuji}.} 
    The work of~\cite{Shen2020Generating} presents multi-objective scatter plots and further visualizes the gameplay trajectories of their agents trained for \textit{Justice Online}.

    Besides, comparison-based methods, QD analysis, and other specialized methods are also employed to measure agent diversity.
    Szubert \textit{et al.} \cite{Szubert2015Role} calculate the average Hamming distance between the action sequences of agents as a measure of behavioral diversity.
    Yuda \textit{et al.} present the minimum, maximum, and average cosine similarity of each agent's featured vector across all others to exhibit their behavior diversity \cite{Yuda2021Identification}.
    Canaan \textit{et al.} leverage QD search to generate diverse agents to play Hanabi \cite{Canaan2019Diverse}. Coverage and averaged QD-score are reported and the QD-maps are illustrated. They also proposed intra-run \revised{and cross-run diversity to evaluate their algorithm's performance in finding behaviorally diverse agents.}
    % Justesen \textit{et al.} compared the \textbf{produced combat units} regarding each trained agent to illustrate the diversity among the agent behaviors \cite{Justesen2019Learning}.
    Halina and Guzdial propose to measure the diversity of an agent relative to others \cite{Halina2022Diversitybased} as the percentage of random states that the agent takes a distinct action compared to each other. A higher distinct percentage identifies a more distinct game-playing policy. 
    Gaina \textit{et al.} tune game parameters via an evolutionary algorithm to minimize the performance of default agents and maximize the performance of specialized agents, assuming that higher fitness value indicates greater strategic diversity \revised{of the maps~\cite{Gaina2017Automatic}. This approach contrasts with other game genres where map diversity might play a more central role. Tan \textit{et al.} proposed a moment matching method to diversify game-playing policies for a mini team-based shooter game~\cite{tan2023policy}.}
    
    \revised{Other genres, such as \textit{first-person shoot (FPS)} games, known for their competitive nature, emphasize the \textbf{balance} of in-game resources and player-versus-player interactions. Consequently, the focus of PCG in these games is primarily on ensuring balanced gameplay, \textbf{fairness} in player confrontations, and \textbf{difficulty} rather than on the internal diversity of the maps~\cite{Cardamone2011Evolving, Cachia2021MultiLevel}.} 
    % \wzqtodo{\cite{tan2023policy} describes the work and highlight the policy diversity for cooperative.} 
    The lack of literature specifically addressing the internal diversity of maps in FPS games suggests that this aspect has not been a major focus in the field. This observation opens up opportunities for future research to explore how PCG can enhance the diversity of map design in FPS games, potentially enriching the player experience by offering varied environments and strategic challenges.

    \revised{In contrast to FPS games, genres like puzzle games or role-playing games might focus more on the diversity of puzzle types or narrative branching paths, respectively. These genres emphasize cognitive challenges or player choice, rather than resource balance or competitive fairness. A more comprehensive meta-analysis of how diversity metrics align with different game attributes in the future could offer deeper insights into the relationship between game genre and player engagement.}

\subsection{Gaps}% Existed in Game Scenario Diversity Evaluation}
    % The gaps between industry and research, as well as across different fields, highlight significant challenges in the development and evaluation of game diversity:

    The gaps existed in industry and academia, emerging technologies, cross-disciplinary, and educational contexts underscore substantial challenges in the cultivation and assessment of game diversity:
    \revised{The gap between industry and research may stem from the immediate needs of the gaming industry to adapt to player feedback for commercial success, whereas academic frameworks often have a longer development cycle and focus on theoretical rigor. Similarly, in education, evaluation methods prioritize creativity and computational thinking due to traditional focuses on skill-building and technical achievement, which may leave broader diversity metrics underexplored.}

    \subsubsection{Industry vs. research in game diversity metrics} 
    There's a notable disconnect between the theoretical frameworks developed in academia for measuring game diversity and their practical application within the industry. While academic circles propose sophisticated and novel methods for assessing diversity, the industry's uptake is often swayed by user feedback~\cite{Widiati2020User, Saurik2023Evaluating} and gameplay data analytics by questionnaires~\cite{Chen2007User,Saas2016Discovering, Tekin2022Game}, rather than rigorous academic models. This indicates a need for researchers to rethink the application of diversity metrics in game design, potentially by leveraging player data and feedback to guide the development of more relevant evaluation tools.

    \subsubsection{Large language models as evaluators} 
    With the advancement of technology, AIs are playing an increasingly crucial role in automating the evaluation and enhancement of diversity within game scenarios. According to Gallotta \textit{et al.}, RAGAS is one of the examples utilizing one large language model (LLM) to evaluate generated game content by other LLMs~\cite{Gallotta2024Large}. This presents an opportunity for the creation of real-time evaluation tools and self-regulating mechanisms for measuring diversity, based on player evaluations, to seamlessly integrate diverse content creation into the game development process.

    \subsubsection{Cross-disciplinary gaps} 
    There is a vast gap between different fields of research in how they approach and measure diversity. For instance, research in multi-agent systems and RL might focus on the diversity of strategies and policy, while PCG emphasizes the diversity of game content. Affective modeling considers both content and player experience diversity but tends to categorize diverse psychological reactions into one or a few emotions. Moreover, very few studies utilize biological data to measure game diversity, despite its potential to provide a comprehensive understanding of diversity that includes content, strategy, player behavior, perceived diversity, and physical responses (e.g., exercise intensity or heart rate), especially relevant for VR or movement-based games~\cite{Yoo2017Evaluating,Yannakakis2008Entertainment,Yannakakis2008Entertainmenta,Yannakakis2007Capturing,Yannakakis2006Capturing}. Fields like serious game design and dynamic difficulty adjustment focus more on learning experience and game difficulty~\cite{Mortazavi2024Dynamic, Wouters2011Measuring,Moizer2019approach}. \revised{This gap represents an opportunity for interdisciplinary research that incorporates player behavioral and physiological data to provide a holistic view of game diversity.}

    \subsubsection{Educational evaluation methods}
    Educational evaluation methods in the context of leveraging the whole student-designed games and simulations for learning often emphasize creativity and the development of computational thinking skills. Koh \textit{et al.} have contributed significantly to this area by developing a methodology that bridges computational thinking with science simulations~\cite{Koh2010Automatic,Bennett2013Computing,Koh2011Computing}. Their approach involves the use of cosine similarity and divergence scores (based on Euclidean distance) to compare student-created simulations against a standard sample. This innovative method not only assesses the creativity and technical accuracy of the simulations but also quantifies the extent to which students internalize and apply computational thinking principles. Additionally, techniques like the use of a similarity matrix to analyze students' creations reveal insightful patterns in the application of programming concepts~\cite{Basawapatna2010Visualizing}. \revised{However, they lack a broader analysis of game diversity metrics. Techniques like cosine similarity and divergence scores are used to assess technical accuracy but do not fully capture the diversity of student approaches or how these may foster deeper learning outcomes. Future research should explore more comprehensive methods to evaluate educational game diversity, encompassing both creative and technical dimensions.}
    
    Addressing these gaps requires a concerted effort to develop interdisciplinary approaches that encompass content diversity, strategy diversity, player behavioral diversity, and emotional diversity. This holistic view should ideally extend to measuring biological responses, thereby offering a more comprehensive assessment of game diversity suited for various types of games, including VR and movement-based games. By bridging these divides, researchers and developers can better understand and enhance the multi-dimensional nature of game diversity, ultimately leading to richer, more engaging gaming experiences for a diverse player base.
    % \item Collaboration between researchers, game developers, and players is crucial in refining diversity metrics and ensuring their practical applicability. This collaboration can bridge the gap between theoretical models and real-world game design practices.
    % \item The potential for improved player experiences through the thoughtful application of diversity metrics is immense. This makes the exploration of diversity in games not only an exciting area of research but also a vital aspect of game development in the industry.
    % \end{itemize}

\section{Conclusion\label{sec:conclusion}}

    In conclusion, our survey highlights the critical role of diversity in video game scenarios, directly impacting player engagement, satisfaction, and overall gaming experience. By examining a spectrum of evaluation methods and metrics, we illuminate the nuanced approaches to measuring game scenario diversity, revealing a complex interplay between content creation and player interaction. Our findings underscore the necessity of a unified taxonomy and a structured strategy for choosing diversity measures that can guide both academic research and practical application in game development.

    The gaps identified between industry practices and academic research, alongside the emerging role of AI and LLMs in automating content generation and evaluation, present both challenges and opportunities. Bridging these gaps requires fostering closer collaboration between researchers and game developers, leveraging the strengths of both worlds to create more diverse, engaging, and innovative game experiences.

    Looking forward, the continuous evolution of PCG technologies and methodologies offers a promising avenue for addressing the identified gaps and further advancing the diversity of game scenarios. Future research should focus on developing adaptive and dynamic evaluation metrics that can better capture the multi-dimensional diversity in games. Moreover, exploring underrepresented game genres and incorporating player feedback and biological data into diversity evaluation could provide deeper insights into the subjective experience of diversity in gaming.

    Embracing diversity in game scenarios not only enriches player experiences but also reflects the diverse interests and backgrounds of the global gaming community. As AI continues to play a pivotal role in shaping the future of game automation, concerted efforts toward understanding and enhancing game scenario diversity will be essential in creating more inclusive, engaging, and innovative gaming environments.
    
   % \section*{Acknowledgments}
        \balance
    \bibliographystyle{IEEEtran}
    \bibliography{refs}
    % \printbibliography
    
\newpage    
    \onecolumn
    %\appendix[Formulation of Diversity Metrics]
    \appendices
    \section*{Supplementary Material: Formulation of Diversity Metrics and Methods in Objective Evaluation}
    This supplementary material presents the formulations of diversity metrics and methods used in objective evaluation.
    %For a comprehensive introduction, citations, and discussions on related works regarding these metrics and methods, please refer to Section \ref{sec:metrics} and Section \ref{sec:objective approach} of the main manuscript, respectively.

    For a comprehensive understanding of the notations used throughout this survey, we provide a detailed description of each symbol and its application in the context of evaluating game scenario diversity. These notations form the backbone of the mathematical models and analyses presented in the following sections.

   \noindent \textbf{Notations:} 
    \begin{itemize}
        \item $\S$: A set of scenarios or game levels under evaluation.
        \item $x$, $x'$: Individual game scenarios or components within the set. 
        \item $d$, $d'$: The length of $x$, $x'$, if the scenarios are represented in vector or sequence.
        \item $P$, $Q$: Probability or frequency distributions of considered items in scenarios, such as tile-pattern distribution.
        \item $P_i$, $Q_i$: The probability or frequency of the $i$th item in $P$ and $Q$, respectively.
        \item $\vec(\cdot)$: The featured vector of its argument.
        \item $v_i(\cdot)$: The $i$th entry of $\vec(\cdot)$.
        \item $\delta(\cdot, \cdot)$: A comparison-based metric capturing the extent of dissimilarity or distance between two scenarios.
        \item $\phi(\cdot)$: A non-comparison-based metric summarizing some characteristic of scenarios into a value.
    \end{itemize}

\subsection{Formulations of metrics for objective evaluation}

This section presents the formulations of diversity metrics used in objective evaluation. For a comprehensive introduction, citations, and discussions on related works regarding these metrics, please refer to Section \ref{sec:metrics} of the main manuscript.

\subsubsection{Formulations of comparison-based metrics}
%\begin{itemize}
%\item \textbf{Distance Measures:}
\paragraph{Distance Measures}
    \begin{itemize}
        \item Cosine similarity/distance
        \begin{equation*}
        \delta(x, x') 
        = \frac{\vec(x) \cdot \vec(x')}{\|\vec(x)\| \|\vec(x')\|} 
        = \frac{\sum_{i=1}^d v_i(x) v_i(x')}{\sqrt{\sum_{i=1}^d v_i^2(x)} \sqrt{\sum_{i=1}^d v_i^2(x')}}.
        \label{eq:cos}
        \end{equation*}
    
        \item Manhattan distance
        \begin{equation*}
        \delta(x, x') = \sum_{i=1}^{d} \big| v_i(x) - v_i(x') \big|.
        \label{eq:manha}
        \end{equation*}
        \item Euclidean distance
        \begin{equation*}
            \delta(x, x') = \sqrt{\sum\nolimits_{i=1}^{d} \big| v_i(x) - v_i(x') \big| ^2}.
            \label{eq:euc}
        \end{equation*}
        \item Hamming distance
        \begin{equation*}
            \delta(x,x') = \sum_{i=1}^{d}{\oneif{[v_i(x)\neq v_i(x')]}}.
        \end{equation*}
        \item Compression distance
        \begin{equation*}
            \delta(x, x') = \frac{C(x \oplus x') - \min\{C(x), C(x')\}}{\max\{C(x), C(x')\}}, 
        \end{equation*}
        where $\oplus$ represents string concatenation, $C(\cdot)$ represents the length of its input after compression.
        \item Edit distance
        \begin{equation*}
        \begin{aligned}
        \delta(x,x')
        &=\delta_{d,d'}, \text{ with} \\
        \forall 1 \leq i \leq d, ~~ \delta_{i,0}
        &=\sum_{k=1}^i w_{\text{del}}\left(v_k(x)\right) \\
        \forall 1 \leq j \leq d', ~~ \delta_{0,j}
        &=\sum_{k=1}^{j} w_{\text{ins}}\left(v_k(x')\right),  \\
        \forall 1 \leq i, j \leq d', ~~
        \delta_{i,j}
        &= \begin{cases}
        \delta_{i-1, j-1}, &\text{if } v_i(x) = v_j(x') \\
        \min 
        \begin{cases}
            \delta_{i-1, j}+w_{\text {del }}\left(v_i(x)\right) & \\
            \delta_{i, j-1}+w_{\text {ins }}\left(v_j(x')\right),  \\
            \delta_{i-1, j-1}+w_{\mathrm{sub}}\left(v_i(x), v_j(x')\right)
        \end{cases} 
        & \text{otherwise}
        \end{cases},
        \end{aligned}
        \end{equation*}
        where $w_{\text {del }}$, $w_{\text {ins }}$ and $w_{\mathrm{sub}}$ are predefined weights for deletion, insertion and substitution, repspectively.
        \item Dynamic time warping
        \begin{equation*}
        \begin{aligned}
            \delta(x,x')
            &= \delta_{d,d'}, \text{ with}\\
            \delta_{i, j} 
            &= \Delta\left(v_i(x),v_j(x')\right) + \min\{\delta_{i-1, j}, \delta_{i, j-1} ,\delta_{i-1, j-1}\},
            \end{aligned}
        \end{equation*}
        where $\Delta(\cdot, \cdot)$ indicates a distance measure of two sequence items.
    \end{itemize}
%\item \textbf{Divergence measures:}
\paragraph{Divergence measures}
    \begin{itemize}
    \item Similarity
          % we summarize this to be simply the sum of the number of rows and columns in the generated content that appear in another content set $\mathcal{S}$. 
    \begin{equation*}
        \delta(x|\mathcal{S}) 
        = \sum_{i=1}^{r} \oneif[\exists y \in \mathcal{S}, \vrow_i(x) =\vrow_i(y)] + \sum_{i=1}^{c} \oneif[\exists y \in \mathcal{S}, \vcol_i(x) =\vcol_i(y)],
    \end{equation*}
    where $\vrow_i(x)$ represents the feature value of $i$th row of $x$ and $\vcol_i(x)$ represents the feature value of the $i$th column of $x$. Typically, the feature used in this metric is the count of some kind of elements.
    \item   KL divergence  
    % When the expectation (weighted sum) is calculated using the probability distribution $P$, it is formulated as
    \begin{equation*}
    \begin{aligned}
        \delta_{KL}\big(P \| Q \big) &= \sum_{i=1}^C {P_i \log \frac{P_i}{Q_i} },    
    \end{aligned}
    \end{equation*}
    where $C$ is the total number of categories.
    \item   JS divergence
    \begin{equation*}
    \delta_{JS}(P \| Q) =\frac{1}{2} \left( \delta_{KL}\left(P \Big\| \frac{P + Q}{2}\right) + \delta_{KL}\left(Q \Big\| \frac{P + Q}{2}\right) \right).
    \end{equation*}
        \item Jaccard distance
        \begin{equation*}
        \delta(\mathcal{S}, \mathcal{S}') = 1 - \frac{|\mathcal{S}\cap \mathcal{S}'|}{|\mathcal{S}\cup \mathcal{S}'|}.
        \label{eq:jac}
        \end{equation*}
        \end{itemize}
\paragraph{Measures based on self-comparison}
        \begin{itemize}
        \item Symmetry
        \begin{equation*}
        \delta(x) 
        = \sum_{k=1}^{r/2} \oneif[\vrow_{r/2 - k}(x) =\vrow_{r/2 + k}(x)] 
        + \sum_{k=1}^{c/2} \oneif[\vcol_{c/2 - k}(x) =\vcol_{r/2 + k}(x)],
        \end{equation*}
    where $\vrow_i(x)$ represents the feature value of $i$th row of $x$ and $\vcol_i(x)$ represents the feature value of the $i$th column of $x$, $r$ and $c$ represent the number of rows and the number of columns, respectively.
\end{itemize}

\subsubsection{Formulations of non-comparison-based metrics}
    \begin{itemize}
    \item Linearity
    \begin{equation*}
    \phi(x) = \frac{1}{n} \sum_{i=1}^n |h_i(x) - L_i(x)|,
    \label{eq:lin}
    \end{equation*}
    where $n$ is the number of units, $h_i(x)$ denotes the height of the $i$th unit in the level $x$, and $L_i(x)$ represents the height of the best fit line for $x$'s unit heights derived through linear regression. 
    \item Leniency
    \begin{equation*}
    \phi(x) = \frac{1}{kn} \sum_{i=1}^n \sum_{j=1}^k  \text{score}(\#\text{item}_{i,j}), 
    \label{eq:len}
    \end{equation*}
    where $n$ is the number of units, and $k$ is the count of item types considered, typically associated with the game's difficulty aspects, $\#\text{item}_{i,j}$ is the count of $j$th item in the $i$th column, $\text{score}(\cdot)$ is a scoring function based on the count of items in a column.
    \item  Density
    \begin{equation*}
    \phi(x) = \frac{\#\text{item}}{\#\text{total\_items}},
    \label{eq:den}
    \end{equation*}
    where $\#\text{item}$ and $\#\text{total\_item}$ are the count of the considered items and the count of all types of items, respectively.
    \end{itemize}

\subsection{Formulations of objective evaluation approaches}
This section presents the formulations of objective evaluation approaches. For a comprehensive introduction, citations, and discussions on related works regarding these methods, please refer to Section \ref{sec:objective approach} of the main manuscript.

    \subsubsection{Single-indicator methods}
    We use $X$ to denote a generative scenario distribution to be evaluated (a set of scenarios can be viewed as a uniform distribution over the set).

    \paragraph{Comparison-based methods}

    \begin{itemize}
    \item Average distance
    \begin{equation*}
    D(X) = \E_{x \sim X, x' \sim X} \left[ \delta(x, x') \right],
    \end{equation*}
    % where $x$ and $x'$ are i.d.d. samples from $X$.

    \item Average nearest neighbor distance
    % A general formulation of this diversity metric is
    \begin{equation}
    D(X| \Re) = \E_{x \sim X} \left[\min_{x' \in \Re} \delta(x, x') \right], 
    \label{eq:annd}
    \end{equation}
    where $\Re$ denotes a \textit{reference set}, consisting of representative content samples, which can be the cluster centers selected by some clustering algorithm. 
    \item Relative diversity
    \begin{equation}
        D(x | \mathcal{S}) = \frac{1}{|\mathcal{S}|} \sum_{x' \in \mathcal{S}} \delta(x, x'), 
    \end{equation}
    \item  Novelty score
    \begin{equation}
      D(x|\mathcal{S}) = \frac{1}{k} \sum_{i=1}^k \delta(x, N_j(\mathcal{S})),
    \end{equation}
    where $N_j(\mathcal{S})$ is the $j$th nearest neighbor in $\mathcal{S}$ in terms of $\delta$, $k$ is a parameter.

    \item Computational surprise
    \begin{equation}
        D(x|\mathcal{S}) = \frac{1}{k} \sum_{j=0}^k \delta(x, M_j(\mathcal{S})),
    \end{equation}
    where $M$ is a model that predicts multiple possible behaviors based on the history, and $M_j(\mathcal{S})$ indicates the $j$-closest prediction to $x$, $k$ is a parameter.

    \item  Slacking A-clipped function 
    % is a method evaluating the diversity of content within a scenario. It is devised for assessing online generated level segments, rewarding the moderate divergences of a new level segment regarding several previously generated ones. It is formulated as
    \begin{equation}
    \begin{aligned}
    D(x_i|x_{i-1}:x_{i-n}) &=  \frac{1}{R} \sum_{k=1}^n \min\left\{1 - \frac{|g-\delta(x_i, x_{i-k})|}{g}, r_k \right\}, ~\text{with}\\
    r_k &= 1-\frac{k}{n+1}, \\
    R &= \sum\nolimits_{k=1}^n r_k,
    \end{aligned}
    \end{equation}
    where $n$ and $g$ are hyperparameters, representing the number of previous segments to be considered and the desired value of divergence, respectively. In this metric, segments are sequential while $x_i$ indicates the $i$th segment in the sequence.
    \end{itemize}

\paragraph{Non-comparison-based methods}
    \begin{itemize}
    \item  Standard deviation
    \begin{equation}
        D(X) = \sqrt{\E\left[\Big( \phi(x) - \E[\phi(x)] \Big)^2\right]},
    \end{equation}  
    \item Behavior diversity
    \begin{equation}
    \begin{aligned}
        D(X) &= \left( \frac{ \sigma_\phi }{\sigma_{\text{max}} } \right)^p, ~ \text{with} \\
        \sigma_{\text{max}} &= \frac{1}{2} \sqrt{\frac{N}{N-1}} (\phi_{\text{max}}(X) - \phi_{\text{min}}(X)),        
    \end{aligned}
    \end{equation}
    where $\sigma_\phi$ is the standard deviation over the $N$ games regarding feature $\phi$, $\phi_{\text{max}(X)}$ and $\phi_{\text{min}(X)}$ are the maximum and minimum values of $\phi$ across $X$, and $p$ is a weighting parameter.

    \item Spatial diversity splits a level or map into some cells, and calculates the entropy over those cells as
    \begin{equation}
    D(x) = - \frac{1}{\log C}\sum_{i=1}^{C} \frac{\phi(x_i)}{\phi(x)} \log \frac{\phi(x_i)}{\phi(x)},
    \end{equation}
    where $C$ is the total number of cells, $x_i$ is the $i$th cell. Note the input $x$ is lowercase because spatial diversity measures the intra-diversity of a single scenario. 
    Notably, $\frac{\phi(x_i)}{\phi(x)}$ should be a frequency, i.e., $\sum_{i=1}^C \phi(x_i) = \phi(x)$ must be hold.
    \item Simpson index
    \begin{equation}
        D(x) = 1 - \sum_{i=1}^C \left( \frac{\phi(x_i)}{\phi(x)} \right)^2,
    \end{equation}
    where $C$ is the number of categories and $\frac{\phi(x_i)}{\phi(x)}$ should be a frequency, similar to spatial diversity.
    \end{itemize}

\subsubsection{Multi-indicator methods}

    ERA and QD analysis often rely on a partition $\Pi$ of the feature (behavior) space $\mathcal{F}$, as defined by metrics in Section \ref{sec:metrics}. A partition of $\F$ is a collection of subsets of $\F$ such that the union of all items in $\Pi$ equals to $\F$, and any two items $\pi$ and $\pi'$ are non-overlapping, i.e., $\pi \cap \pi' = \varnothing$. In the context of multi-indicator approaches, the evaluation involves a set of samples $\S$ indexed by the partition $\Pi$. We denote the samples within $\S$ corresponding to a specific partition item $\pi$ as $\S(\pi)$, expressed as $\S(\pi) = \{x | x \in \S \land \f(x) \in \pi\}$. It's worth noting that the $\Pi$ can be a non-uniform partition.
    \begin{itemize}
        \item QD score  \begin{equation}
    \mathrm{QDS}(\S) = \frac{1}{|\Pi|} \sum_{\pi \in \Pi} \max_{x \in \S(\pi)} q(x),
    \end{equation}
    where $q(x)$ is the quality score of $x$.

    \item Coverage \begin{equation}
    \mathrm{Cov}(S) = \frac{1}{|\Pi|} \sum_{\pi \in \Pi} \oneif[\S(\pi) \neq \varnothing].
    \end{equation}
    \item Hypervolume
    \begin{equation}
        \mathrm{HV}(X|\mathbf{r}) = \Lambda \left( \bigcup\nolimits_{x \in X} \{p \in \mathbb{R}^m ~|~  p \succ \mathbf{r} \land \vec(x) \succ p\} \right),
    \end{equation}
    where $\Lambda(\cdot)$ indicates the Lebesgue measure, $m$ is the number of objectives, $\succ$ indicates dominating, $\vec(x)$ is the vector of $x$'s all objective values, and $\mathbf{r}$ is a $m$-dimensional reference point.
    
    \end{itemize}
%Notations:
%一句话说明这里formulate的metrics是正文第几章的内容，说明介绍、citation和相关工作见正文（这里再cite一遍也没毛病）

%另起一段话介绍notation，可以是列表或表格

%列表或表格式介绍formulation

    \end{document}